\documentclass{article}

\usepackage[preprint]{corl_2026}

\usepackage{microtype}
\usepackage{graphicx}
\usepackage{subcaption}
\usepackage{booktabs}
\usepackage[table]{xcolor}
\usepackage{amsmath}
\usepackage{amssymb}
\usepackage{mathtools}
\usepackage{amsthm}
\usepackage{bm}
\usepackage{bbm}
\usepackage{mathrsfs}
\usepackage[capitalize,noabbrev]{cleveref}
\usepackage{placeins}

\newcommand{\rwname}{\texttt{TOPReward}}
\newcommand{\dname}{\texttt{Mani\allowbreak Reward\allowbreak Bench}}
\newcommand{\Norm}[1]{\left\| #1 \right\|}
\newcommand{\set}[1]{\left\{ #1 \right\}}

\def\beq{\begin{equation}}
\def\endeq{\end{equation}}

\newcommand{\Be}{\begin{equation}}
\newcommand{\Ee}{\end{equation}}

\newcommand{\Bm}{\begin{multline}}
\newcommand{\Em}{\end{multline}}
\newcommand{\Bea}{\begin{eqnarray}}
\newcommand{\Eea}{\end{eqnarray}}
\newcommand{\Beas}{\begin{eqnarray*}}
	\newcommand{\Eeas}{\end{eqnarray*}}
\newcommand{\Benu}{\begin{enumerate}}
	\newcommand{\Eenu}{\end{enumerate}}
\newcommand{\Bi}{\begin{itemize}}
	\newcommand{\Ei}{\end{itemize}}

\def\intslash{\rlap{\kern  .32em $\mspace {.5mu}\backslash$ }\int}
\def\qsl{{\rlap{\kern  .32em $\mspace {.5mu}\backslash$ }\int_{Q_x}}}

\def\emph#1{{\it #1 }}

\def\be#1{\begin{equation}\label{ #1}}
\def\endeq{\end{equation}}
\def\endal{\end{align}}
\def\bas{\begin{align*}}
\def\eas{\end{align*}}
\def\bi{\begin{itemize}}
\def\ei{\end{itemize}}

\def\emph#1{{\it #1}}
\def\textbf#1{{\bf #1}}

\def\inf{\infty}

\theoremstyle{plain}

\theoremstyle{definition}

\theoremstyle{remark}

\title{TOPReward: Token Probabilities as Hidden Zero-Shot Rewards for Robotics}

\author{%
  Shirui Chen$^{1,2}$ \And
  Cole Harrison$^{3}$ \And
  Ying-Chun Lee$^{1}$ \AND
  Angela Jin Yang$^{1}$ \And
  Zhongzheng Ren$^{1,2,4}$ \And
  Lillian J. Ratliff$^{1}$ \AND
  Jiafei Duan$^{\dagger,1,2}$ \And
  Dieter Fox$^{\dagger,1,2}$ \And
  Ranjay Krishna$^{\dagger,1,2}$
}

\AtBeginDocument{%
  \hypersetup{%
    pdftitle={TOPReward: Token Probabilities as Hidden Zero-Shot Rewards for Robotics},
    pdfauthor={Shirui Chen, Cole Harrison, Ying-Chun Lee, Angela Jin Yang, Zhongzheng Ren, Lillian J. Ratliff, Jiafei Duan, Dieter Fox, Ranjay Krishna},
    pdfsubject={Preprint}%
  }%
}

\begin{document}
\maketitle
\begingroup
\renewcommand{\thefootnote}{}
\footnotetext{%
  $^\dagger$Co-advised.
  $^1$University of Washington
  $^2$Allen Institute for AI
  $^3$Amazon
  $^4$University of North Carolina at Chapel Hill.
  Correspondence to: Shirui Chen
  \href{mailto:sc256@uw.edu}{\texttt{<sc256@uw.edu>}}.
}
\addtocounter{footnote}{-1}
\endgroup
\vspace{-0.25in}
{\centering\small
  \url{https://topreward.github.io/webpage/}
  \par
}
\vspace{0.15in}

\begin{abstract}
General-purpose robot learning requires dense, instruction-conditioned feedback that can distinguish meaningful task progress from stalled, failed, or partially completed behavior. Yet obtaining such feedback at scale remains difficult, since existing approaches often rely on manual progress annotations, task-specific demonstrations, or reward models trained on curated robot datasets. We introduce $\rwname$, a training-free progress reward method that probes pretrained Video-Language Models (VLMs) through their internal token probabilities rather than asking them to generate numerical progress values. Given a video prefix and a language instruction, $\rwname$ measures the model's likelihood that the instructed task has been completed, converting latent video-language understanding into a dense reward signal without task-specific reward-model training or manually annotated progress labels. We evaluate $\rwname$ on ManiRewardBench, our real-world manipulation benchmark spanning 130 unique tasks and four robot platforms, as well as on Open X-Embodiment datasets. Across these settings, $\rwname$ substantially outperforms prior training-free VLM reward methods on open-source models and is competitive with a trained reward-model baseline on progress-estimation metrics, while requiring no reward-model training. Additional analyses show that the reward is sensitive to the specified instruction and is not explained by time index alone. Finally, $\rwname$ supports downstream applications including success detection and offline reward-weighted behavior cloning.
\end{abstract}


\section{Introduction}
\label{sec:introduction}
Recent advances in Vision-Language-Action (VLA) models have spurred significant interest in leveraging Reinforcement Learning (RL) to achieve truly generalizable real-world performance~\citet{luo2025rl100,chen2025sarm,xiao2025self}. However, real-world RL remains bottlenecked by the extreme sample inefficiency inherent in sparse reward signals. To bridge this gap, the research community has pivoted toward developing generalizable process reward models that provide fine-grained and dense feedback. Current efforts typically focus on directly fine-tuning vision-language models (VLMs) as process reward functions on curated robot datasets~\citet{duan2024aha,budzianowski2025opengvl,ma2024vision,lin2025failsafe} or training specific networks with custom-collected datasets~\cite{zhang2025rewind,chen2025sarm}. For instance, RoboDopamine~\citep{tan2025robo} trains a reward model on 3,400+ hours of manipulation data with step-aware multi-view perception, but requires task-specific demonstrations for adaptation. Likewise,~\citet{lee2026roboreward} introduces RoboReward, which fine-tunes a VLM on a large-scale set of robot trajectories with human-provided success labels and progress scores. However, these approaches rely on costly data requirements: RoboDopamine requires additional demonstrations when adapting to each new task, and RoboReward reports clear gaps across different embodiments and views, indicating limited generalization guarantees. Therefore, while these efforts demonstrate promise for narrow-domain reward models, they still rely on extensive data collection and struggle to generalize beyond the training distribution.

To circumvent the high costs of task-specific fine-tuning, we investigate the use of pretrained VLMs as zero-shot reward models. Our goal is to harness the pretrained visual-language representations embedded in these models to provide generalizable instruction-conditioned progress signals. Recent literature~\citep{rocamonde2023vision, ma2024vision, baumli2023vision} has converged on progress estimation as a useful proxy for value, as it provides dense temporal feedback for learning and adaptation. The current state-of-the-art training-free method, GVL~\citep{ma2024vision}, casts progress estimation as visual question-answering—but performs well only
on proprietary VLMs like Gemini and GPT-4~\citep{budzianowski2025opengvl}, collapsing on open-source alternatives. Indeed, contemporary studies~\citep{zhang2026progresslm} suggest that open-source VLMs are not yet ``robotics-ready'' for progress estimation.

\begin{figure*}[t]
    \centering
    \includegraphics[width=\textwidth]{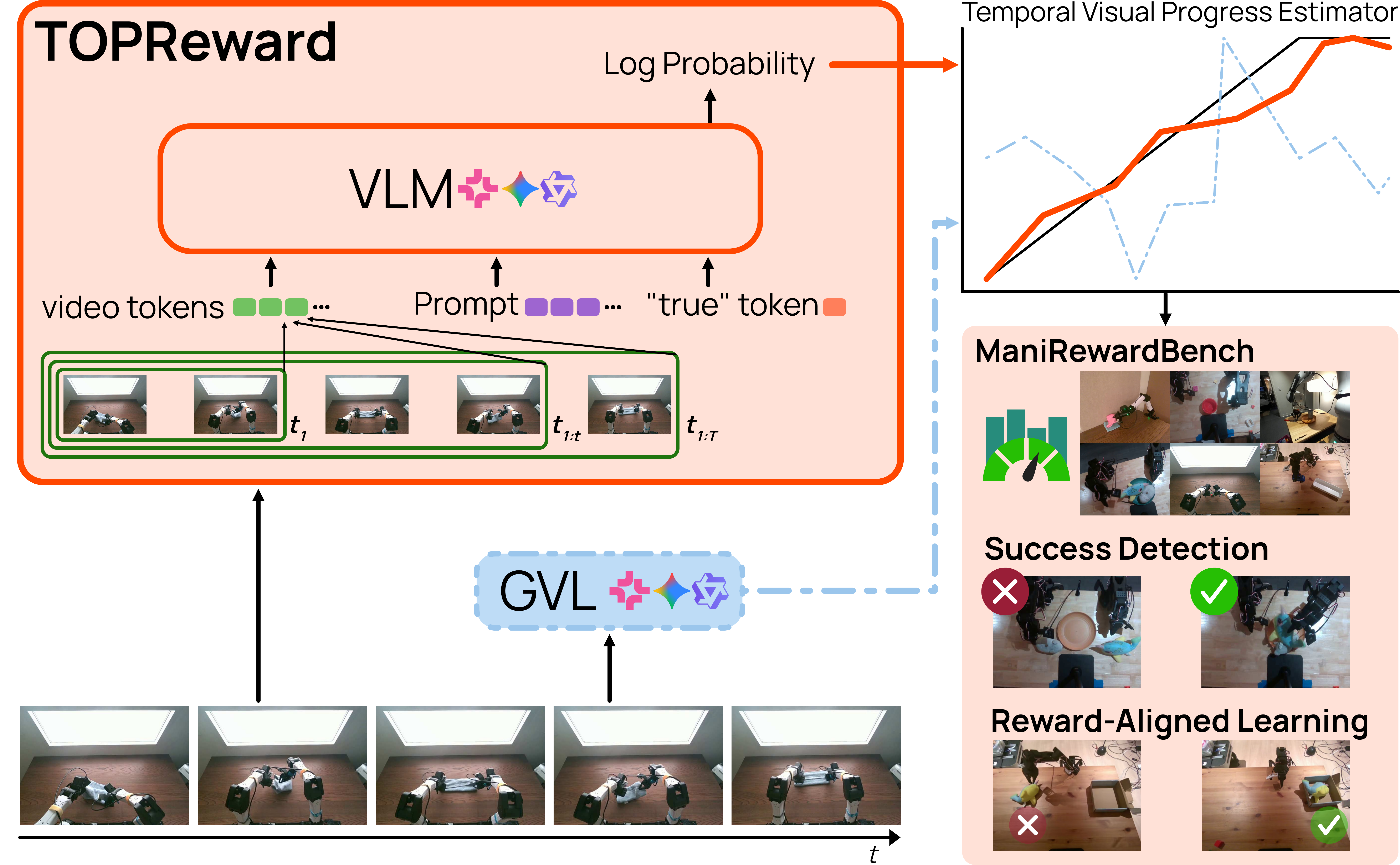}
    \caption{\textbf{Result highlights}. $\texttt{\rwname}$ enables effective zero-shot estimation of task progress across diverse and challenging real-world manipulation tasks, without task-specific training. Across several VLM backbones, $\texttt{\rwname}$ provides a temporally consistent visual reward signal that supports multiple downstream applications, including success detection, policy improvement, and evaluation on our in-house benchmark, $\dname$.}
    \label{fig:main_results}
\end{figure*}
In this work, we challenge the prevailing assumption that open-source VLMs are unfit for reward modeling by introducing a novel formulation for zero-shot progress estimation. We hypothesize that the failure of current open-source models stems not from a lack of temporal understanding, but from the representation bottleneck of textual output—specifically, the models' inconsistent instruction-following and their notorious bias in representing numerical tokens. To resolve this, we present $\texttt{\rwname}$, a probabilistically grounded progress estimator that bypasses autoregressive text generation entirely. Instead of prompting the VLM to output the value of the completion percentage in text space, $\texttt{\rwname}$ extracts the model's internal ``belief'' by analyzing the probabilistic distribution of its token logits. By measuring the shift in confidence toward task-completion tokens over time, we derive a continuous, well-posed progress signal directly from the VLM's latent probabilistic representation. This approach requires zero additional training or fine-tuning, revealing that robust reward modeling is an emergent capability already present in pretrained video VLMs—if one looks beyond the text.

To enable rigorous evaluation of progress estimation methods, we introduce $\dname$, a benchmark comprising 130 unique real-world manipulation tasks spanning multiple robot platforms (Franka Emika arms, Single-Arm/Bimanual YAM, SO-100/101) with temporal annotations of task progress.
We demonstrate that $\texttt{\rwname}$ can effectively track task progress across this diverse benchmark. The raw completion probability supports cross-trajectory success detection, while normalized prefix scores provide within-trajectory progress curves for evaluation and visualization. We further use $\texttt{\rwname}$ to weight expert examples in offline behavior cloning. In real-world deployment on six single-arm SO-100 manipulation tasks, reward-weighted fine-tuning with $\texttt{\rwname}$ consistently improves success rates over standard behavior cloning, achieving up to 10 out of 10 successes on challenging tasks where baseline behavior cloning reaches only 7 out of 10.

\section{Related Work}

\textbf{The reward bottleneck for VLA.}
Large-scale vision-language-action policies such as OpenVLA~\cite{kim2024openvla}, $\pi_0$~\cite{black2024pi0}, MolmoAct~\cite{lee2025molmoact} and Gemini Robotics~\cite{team2025gemini_robotics} have demonstrated strong language-conditioned manipulation capabilities across diverse embodiments, yet reliably deploying them in real-world settings remains an open problem~\cite{firoozi2025foundation}. A natural path forward is reinforcement learning with online or offline fine-tuning, but RL in practice hinges on the availability of a reward signal---one that is traditionally hand-crafted per task in robotics, difficult to scale, and brittle under distribution shift~\cite{kober2013reinforcement, dulac2019challenges}. Recent efforts have applied RL to improve generalist robot policies in real-world deployment~\cite{zhang2024, nakamoto2024, chen2025b, hu2025, ankile2025, wagenmaker2025, dong2025}, including RL-100~\cite{luo2025rl100}, which trains diffusion-based visuomotor policies directly on real robots using human-provided success signals, and $\pi^*_{0.6}$~\cite{black2025pistar}, which improves $\pi_0$ through real-world RL with human-annotated episode outcomes. All of these approaches, however, remain reliant on manual reward specification, underscoring the need for automated, scalable alternatives.

\textbf{Learned reward models.}
A long line of work seeks to replace hand-crafted rewards with learned alternatives~\cite{pomerleau1991efficient, ho2016generative, Hester2017LearningFD}. Embedding-based methods such as VIP~\cite{ma2022vip}, LIV~\cite{ma2023liv}, and R3M~\cite{nair2022r3m} learn visual representations that capture progress toward a goal, but require task-specific fine-tuning and offer limited language grounding. VQA-style approaches such as SuccessVQA~\cite{du2023vision} and related frameworks~\cite{stone2023open, huang2022inner} reframe reward as a binary classification problem---asking a VLM whether the task succeeded---but produce signals too coarse for dense reward shaping~\cite{lynch2020learning, andrychowicz2017hindsight}. Code-generation strategies like Eureka~\cite{ma2023eureka} synthesize reward functions via LLMs, yet depend on access to simulation ground truth. More recently, generalist reward models such as RoboReward~\cite{lee2026roboreward} and RoboDopamine~\cite{tan2025robo} are trained on large-scale datasets of successes and failures to predict progress scores or distance-to-goal estimates~\cite{wu2023unleashing, fan2022minedojo}. While these models move toward broader coverage, they still require domain-specific training data and can struggle to generalize across embodiments and environments~\cite{kalashnikov2021mt}. A common thread across all these approaches is the dependence on training or domain-specific resources---a requirement our method avoids.

\textbf{Training-free value estimation with VLMs.}
A separate and more relevant line of work asks whether VLMs can estimate task progress without any additional training. Generative Value Learning (GVL)~\cite{ma2024vision} poses progress prediction as a temporal ordering problem: given a batch of shuffled trajectory frames, the VLM is prompted to assign per-frame progress scores, exploiting its semantic grounding to rank frames by task completion. This enables various downstream applications such as dataset filtering and advantage-weighted regression without task-specific reward engineering. The OpenGVL benchmark~\cite{budzianowski2025opengvl} evaluates this paradigm across diverse tasks and model families, revealing that open-source VLMs fall substantially short of their proprietary counterparts on temporal progress prediction. In this paper, we hypothesize that the reason for this is not a lack of visual understanding in open-source VLMs but rather the instability of numeric token generation---LLMs are poorly calibrated when asked to produce precise numerical outputs~\cite{wallace2019nlp, yuan2023well}. This observation motivates a fundamental shift: rather than asking a VLM to \emph{generate} a progress value, one can instead probe what the model already \emph{knows} through its internal representations.

\textbf{Internal representations as reward signals.}
A growing body of work in NLP has shown that a model's internal activations---logits, hidden states, and embeddings---track its certainty and factual accuracy more reliably than its generated text~\cite{kadavath2022language, tian2023just, azaria2023internal, burns2022discovering, liu2023cognitive}. In robotics, recent methods have begun to leverage such representations for reward definition~\cite{rocamonde2023vision, grislain2025failsense}, bypassing the instabilities of text generation. Our work, $\rwname$, takes this principle further: instead of prompting a VLM to generate numeric progress estimates, we pose a binary completion query (``does this trajectory complete the task?'') and extract the probability of the affirmative token as a continuous reward signal. This formulation is zero-shot, requires no fine-tuning or domain-specific data, and yields a useful progress signal that scales to the 130-task $\dname$ benchmark as well as the Open X-Embodiment dataset across multiple robot platforms.

\section{\rwname}
\label{sec:method}
The stark difference between GVL's~\citep{ma2024vision} performance on Gemini versus open-source models might mislead one into believing that only the most powerful VLM can accurately estimate the progress of a robotic trajectory. Indeed, for a model to output well-formatted and accurate progress estimates for all shuffled frames, it needs strong instruction-following capability and an accurate internal representation of numerical values. However, neither of these is necessarily correlated with the model's underlying video understanding capability. Therefore, we propose \texttt{\rwname}, a method that leverages the internal understanding of the VLM to produce a progress estimator without requiring accurate, well-formatted numerical generation.

\noindent\textbf{Problem setup.} We formulate the progress estimation task as follows: given an instruction $x$ and a video trajectory $\tau_{1:T}=(I_1,\dots,I_T)$ (frames in chronological order), our goal is to produce a scalar prefix score for each $\tau_{1:t}$ that reflects accumulated evidence that the instruction has been completed.

\subsection{Token probability as the reward} \label{sec:probreward}

\noindent\textbf{Key idea.}
% Rather than prompting a video VLM to \emph{say} a percentage (which requires reliable numerical generation and parsing),
We use the VLM's \emph{internal output, i.e., predicted token probabilities} as the reward. Concretely, we ask the model to judge whether the observed trajectory \emph{completes} the instruction and score the probability of an affirmative answer (e.g. the token \texttt{True}).

% \paragraph{Token log-likelihood as the reward.}
Let $p_\theta$ be a VLM defining a next-token distribution with pretrained weights $\theta$. We form a prompt $u$ that grounds the judgment in the video:
\begin{quote}\small
``\texttt{<|video|>} The above video shows a robot manipulation trajectory that completes the following task: \{\texttt{INSTRUCTION}\}. Decide whether the above statement is True or not. The answer is: $\set{a}$''
\end{quote}
and compute the probability of the answer token sequence $a$=``True'' \footnote{We also explore another variant that evaluates the probability over the entire instruction which consists of multiple tokens, but found it to be less effective. See~\Cref{appendix:alternative_reward} for details.}. We choose \texttt{True} because we found boolean tokens to show the clearest success--failure separation, with \texttt{True} exhibiting the largest absolute difference in mean token probability across episodes (see~\Cref{appendix:true_token} for the token-probability comparison). Denoting the (video-conditioned) textual context by $c(\tau_{1:t},u)$, we define the reward for a prefix to be
\begin{equation}
r_t \;=\; \log p_\theta\!\left(a \mid c(\tau_{1:t},u)\right).
\label{eq:instr_reward}
\end{equation}
In this way, we construct a conditional completion score for the trajectory-instruction pair while sidestepping the need for the language model to generate calibrated numerical values. The raw score $r_t$ is causal because it depends only on the prefix $\tau_{1:t}$ and the instruction. As we will see in~\Cref{sec:experiment}, $r_t$ tends to increase on successful demonstrations as visual evidence accumulates, while its absolute level is useful for comparing complete successful and failed trajectories.

\paragraph{Chat templates.} For open-source \rwname\ experiments, we score the prompt directly rather than adding a generation-oriented chat template. Our ablation in~\Cref{app:chat_ablation} shows that chat templates can substantially reduce performance, so we use direct prompt scoring whenever the model interface permits it. The Gemini API enforces chat-style formatting, which may partly explain its weaker \rwname\ results in~\Cref{tab:oxe_voc}.

\subsection{Progress estimation from trajectory prefixes}

\noindent\textbf{Prefix sampling.}
To obtain a temporal progress curve, we evaluate Eq.~\eqref{eq:instr_reward} on a set of $K$ uniformly spaced prefix lengths $\{t_k\}_{k=1}^K$ with $1 = t_1 < \dots < t_K = T$. This involves $K$ model forwards and produces rewards $\{r_{t_k}\}$ summarizing how completion evidence accumulates over time (\Cref{fig:method_example}).

\noindent\textbf{Normalization.}
The raw log-probability range is $(-\inf, 0]$, while VOC and visualizations are easier to interpret on a bounded scale. For within-trajectory evaluation, we therefore use min-max normalization to map rewards to a normalized progress score $s_{t_k}$ within each episode:
\begin{equation}
s_{t_k} \;=\; \frac{r_{t_k} - \min_j r_{t_j}}{\max_j r_{t_j}- \min_j r_{t_j} + \varepsilon},
\label{eq:minmax}
\end{equation}
with a small $\varepsilon$ for numerical stability. This normalization is non-causal because the episode-level minimum and maximum are known only after all sampled prefixes are scored. We use it for within-trajectory progress visualization and VOC evaluation, not as the raw online reward signal.

\noindent\textbf{Positive weights for downstream offline learning.}
When a per-step weight is needed for reward-weighted offline behavior cloning, we use the \rwname increment to construct a positive weighting signal:
\begin{equation} \label{eq:adv}
\Delta_{t_k} \;=\; \min\left\{ \exp(\beta\cdot(r_{t_k} - r_{t_{k-1}})), \delta_{\max}\right\},
\end{equation}
where $\beta$ controls the sharpness of the weighting and $\delta_{\max}$ caps large weights for training stability. The weight is always positive: steps with decreasing \rwname\ score are downweighted rather than treated as negative supervision, which keeps the flow-matching objective well-posed.
% The final advantage-weighted BC loss is then computed as
% \begin{equation}
% \mathcal{L}(\theta) \;=\; - \mathbb{E}_{(s_t,a_t)\sim\mathcal{D}} \left[ \Delta_t \cdot \log \pi_\tet(a_t|s_t) \right].
% \end{equation}

\section{Experiments}
\label{sec:experiment}

We evaluate $\texttt{\rwname}$ across three main dimensions: (1) zero-shot progress estimation on large-scale robot datasets using VOC and complementary progress metrics (\Cref{sec:zeroshot_eval}); (2) instruction sensitivity; and (3) downstream applications including success detection and real-world reward-weighted behavior cloning.

\noindent\textbf{VLM backbones.}
We evaluate $\texttt{\rwname}$ on Qwen3-VL-8B and Qwen3-VL-32B~\citep{bai2025qwen3vltechnicalreport}, Molmo2-8B~\citep{clark2026molmo2openweightsdata}, and Gemini-2.5-Pro~\citep{comanici2025gemini25pushingfrontier}. Qwen3-VL and Molmo2 are open-source video-language models; Gemini-2.5-Pro serves as a proprietary baseline with logit access.

\noindent\textbf{Benchmark.}
\dname\ contains 130 unique real-world manipulation tasks collected across four robot platforms: Franka, SO-100/101, single-arm YAM, and bimanual YAM. Episodes are annotated with subtask boundaries for stage-aware progress evaluation, and a separate 23-task failure split supports success detection. Additional dataset details are in~\Cref{app:add_bench}.

\subsection{Large-scale real-world evaluation}
\label{sec:zeroshot_eval}
To evaluate zero-shot progress estimation, we test whether \rwname produces accurate dense progress estimates on expert robotic trajectories from \dname\ and Open X-Embodiment (OXE). We mainly compare against GVL~\citep{budzianowski2025opengvl,ma2024vision}, the state-of-the-art training-free progress estimator, which prompts a VLM to assign numerical progress values to shuffled frames.

\noindent\textbf{Metrics.}
Following standard practice~\citep{ma2024vision,ma2023liv}, we use Value-Order Correlation (VOC) to measure Spearman's rank correlation between the chronological order of input video frames and the predicted values,
\begin{equation}
    \begin{split}
        \text{VOC} = \text{rank-correlation}\bigl(\text{argsort}(s_{t_1}, s_{t_2}, \cdots, s_{t_K}), (t_1, t_2, \cdots, t_K)\bigr).
    \end{split}
\end{equation}
VOC ranges from $-1$ to $1$. When VOC equals $-1$, the predicted order is exactly the opposite of the ground truth, and VOC $=1$ indicates perfect alignment. Because VOC measures only temporal ordering, it should not be interpreted as a standalone test of instruction grounding. We therefore report complementary progress metrics (Kendall tau-b, Pearson correlation, and MAE) against the available completion annotations.

\noindent\textbf{Results on Open X-Embodiment.} The Open X-Embodiment (OXE) dataset~\citep{o2024open} is a collection of 50 academic robot datasets spanning diverse tasks, camera configurations, and robot platforms. We select 33 high-quality OXE datasets from the LeRobot collection and use this same filtered set for every row in~\Cref{tab:oxe_voc}. On OXE datasets, \rwname\ substantially outperforms GVL on open-source models such as Qwen3-VL and Molmo. Complementary progress metrics show the same trend beyond VOC across available rows. We also compare against the trained Robometer-4B progress head on the same filtered OXE split using 15 uniformly sampled frames per episode, and find that \rwname is slightly better. On the proprietary Gemini-2.5-Pro, GVL performs better (0.572) while \rwname\ achieves 0.430, reflecting the chat-template issue discussed in~\Cref{app:chat_ablation}.

\begin{table}[t]
\centering
\footnotesize
\setlength{\tabcolsep}{3pt}
\caption{\textbf{Results on the filtered Open X-Embodiment dataset.} VOC reports mean dataset-level VOC over the same 33 filtered datasets and 20 episodes per dataset. Higher is better except MAE.}
\begin{tabular}{llcccc}
\toprule
\textbf{Method} & \textbf{Model} & \textbf{VOC} & \textbf{Kendall} & \textbf{Pearson} & \textbf{MAE} \\
\midrule
GVL & Molmo2-8B & -0.019 & 0.000 & 0.001 & 0.500 \\
GVL & Qwen3-VL-8B & 0.218 & 0.192 & 0.443 & 0.396 \\
GVL & Gemini-2.5-Pro & 0.572 & 0.534 & 0.669 & 0.218 \\
\rwname & Molmo2-8B & 0.525 & 0.427 & 0.503 & 0.308 \\
\rwname & Qwen3-VL-8B & 0.874 & 0.771 & 0.856 & \textbf{0.163} \\
\rwname & Qwen3-VL-32B & \textbf{0.890} & \textbf{0.799} & \textbf{0.870} & 0.191 \\
\rwname & Gemini-2.5-Pro & 0.430 & 0.332 & 0.399 & 0.321 \\
Robometer & 4B & 0.838 & 0.765 & 0.840 & 0.180 \\
\bottomrule
\end{tabular}
\label{tab:oxe_voc}
% \vspace{-0.2cm}
\end{table}

\begin{table}[t]
\centering
\footnotesize
\setlength{\tabcolsep}{3pt}
\caption{\textbf{Results on \dname.} We report metrics averaged over the 497 successful-evaluation episodes. This progress-evaluation subset contains 113 tasks across four robot platforms: LeRobot, Franka, Bimanual YAM, and Single-arm YAM. VOC and Kendall tau-b measure rank agreement, Pearson measures linear agreement, and MAE measures normalized progress error. Higher is better except MAE.}
\begin{tabular}{llcccc}
\toprule
\textbf{Method} & \textbf{Model} & \textbf{VOC} & \textbf{Kendall} & \textbf{Pearson} & \textbf{MAE} \\
\midrule
GVL & Molmo2-8B & -0.003 & -0.002 & -0.003 & 0.514 \\
GVL & Qwen3-VL-8B & 0.316 & 0.261 & 0.326 & 0.392 \\
\rwname & Molmo2-8B & 0.619 & 0.484 & 0.569 & 0.316 \\
\rwname & Qwen3-VL-8B & \textbf{0.942} & \textbf{0.858} & \textbf{0.915} & \textbf{0.126} \\
\rwname & Qwen3-VL-32B & 0.864 & 0.771 & 0.824 & 0.194 \\
Robometer & 4B & 0.858 & 0.768 & 0.861 & 0.160 \\
\bottomrule
\end{tabular}
\label{tab:bench_voc}
\end{table}

\noindent\textbf{Results on \dname.}
We evaluate progress estimation on the successful-trajectory subset of \dname, covering 113 tasks and 497 episodes across 4 robotic platforms. Results are shown in~\Cref{tab:bench_voc}. On Qwen3-VL-8B, \rwname\ achieves the strongest aggregate progress metrics, substantially outperforming GVL on the same backbone. On Molmo2-8B, GVL is near zero across rank and correlation metrics, while \rwname\ recovers a useful progress signal with positive VOC, Kendall tau-b, and Pearson correlation. Compared with the trained Robometer-4B baseline, \rwname\ is competitive across rank, correlation, and error metrics while requiring no reward-model training. Detailed per-task breakdowns and distribution plots are in~\Cref{app:dataset_breakdown}.

\noindent\textbf{Qualitative results.} We visualize representative progress traces in \Cref{fig:linear_progress_examples}. The traces demonstrate that \rwname\ produces smooth, monotonically increasing progress signals that closely track stage-aware ground-truth task completion (computed from annotated subtask boundaries) across diverse manipulation tasks. In contrast, Gemini-GVL exhibits noisier predictions with frequent non-monotonic fluctuations. Notably, \rwname\ correctly captures the temporal structure of multi-step tasks, with progress plateaus corresponding to intermediate subtask completions and accelerations during active manipulation phases.

\subsection{Instruction sensitivity}
\label{sec:instruction_sensitivity}
\begin{figure}[t]
    \centering
    \includegraphics[width=0.99\linewidth]{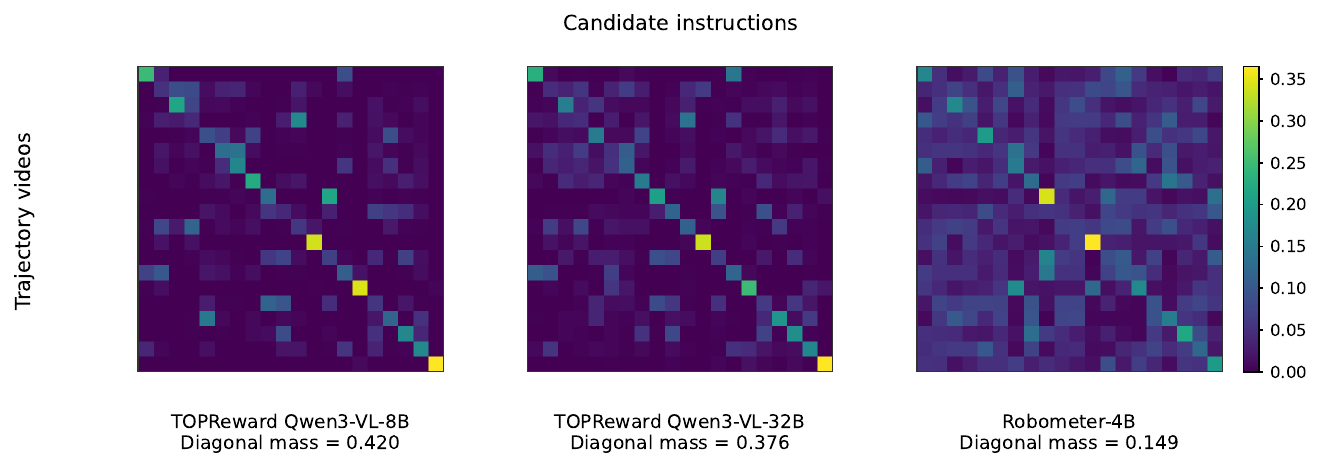}
    \caption{\textbf{Instruction sensitivity.} Instruction-grounding matrices for TOPReward and Robometer-4B. Rows are terminal trajectory videos and columns are candidate task instructions. For TOPReward, each cell is the terminal TOPReward score for that video--instruction pair; for Robometer-4B, each cell is its terminal reward prediction. We apply Sinkhorn normalization for visualization and report mean diagonal mass, so both rows and columns sum to one and bright diagonal cells indicate strong one-to-one matching between trajectories and their intended instructions.}
    \label{fig:instruction_grounding}
\end{figure}

Order-based metrics alone cannot establish that a reward is grounded in the language instruction: an instruction-agnostic ordering score can be highly correlated with progress on successful demonstrations while assigning the same values for every instruction. We therefore evaluate whether final-trajectory rewards discriminate the matched instruction from plausible but incorrect task descriptions. For each video, we score the same terminal trajectory under each candidate instruction and form a video-task $\times$ instruction-task matrix, shown in~\Cref{fig:instruction_grounding}. For TOPReward, each raw entry is the terminal TOPReward score for a video--instruction pair; for Robometer-4B, each raw entry is its terminal reward prediction. We convert these score matrices to positive weights and apply Sinkhorn normalization, so each row and each column sums to one. Strong diagonal structure and high mean diagonal mass in this doubly normalized matrix indicate one-to-one matching between trajectories and their intended instructions rather than instruction-agnostic temporal ordering.

\Cref{fig:instruction_grounding} summarizes instruction disambiguation using mean diagonal mass after Sinkhorn normalization of the displayed score matrix. This gives a bidirectional one-to-one matching view: each trajectory distributes mass across candidate instructions, and each instruction distributes mass across candidate trajectories. The random baseline is $1/20=0.05$ on this 20-task split. The diagonal mass is 0.420 for \rwname\ Qwen3-VL-8B, 0.376 for \rwname\ Qwen3-VL-32B, and 0.149 for Robometer-4B, showing that TOPReward produces substantially stronger matched video--instruction structure than the trained reward-model baseline. As a scalar control on the 150 successful LeRobot episodes in the progress-evaluation subset, pairing each video with its correct instruction gives mean raw reward $-1.68$, while pairing the same videos with mismatched instructions from other tasks drops the mean raw reward to $-5.04$. Thus raw instruction-conditioned rewards and confusion matrices provide the evidence for semantic grounding beyond temporal ordering.

\subsection{Success detection} \label{sec:success_detection}

\noindent\textbf{Setup and results.} We evaluate on the cleaned failure trajectory split of \dname, which contains 150 successful and 129 failed LeRobot attempts. Taking inspiration from self-consistency prompting \citep{self_const}, we combine multiple TOPReward scores using different prompts to obtain a success score for success detection as defined in~\Cref{app:success_score}.

\begin{table}[t]
\centering
\footnotesize
\setlength{\tabcolsep}{4pt}
\caption{\textbf{Success detection results.} We evaluate 279 cleaned LeRobot episodes from \dname\ (150 successful, 129 failed). \rwname\ uses the training-free prefix-margin plus full-video Yes/No score defined in~\Cref{app:success_score}; GVL uses per-trajectory VOC; Robometer-4B uses its trained success head.}
\begin{tabular}{llcc}
\toprule
\textbf{Method} & \textbf{Score} & \textbf{ROC-AUC} & \textbf{PR-AUC} \\
\midrule
GVL (Qwen3-VL-8B) & VOC & 0.728 & 0.893 \\
\rwname\ (Qwen3-VL-8B) & success score & 0.939 & 0.952 \\
\rwname\ (Qwen3-VL-32B) & success score & \textbf{0.965} & \textbf{0.973} \\
Robometer-4B & trained success head & 0.930 & 0.951 \\
\bottomrule
\end{tabular}
\label{tab:success_detection}
\end{table}

As shown in~\Cref{tab:success_detection}, Qwen3-VL-32B reaches 0.965 ROC-AUC and Qwen3-VL-8B reaches 0.939 ROC-AUC, both exceeding the trained Robometer-4B success head without reward-model training or success-label supervision.

\subsection{Real-world reward-weighted behavior cloning}
To further showcase \rwname\ as a signal for policy improvement, we use it to derive positive weights for an offline flow-matching behavior cloning objective. We start from a base policy $\pi_0$ pretrained on 200 hours of the publicly available single-arm SO-100 dataset (HuggingFace). For each of six real-world tasks, we collect 50 demonstrations (potentially noisy and suboptimal) and use \rwname\ to compute prefix scores for the demonstrations. We convert progress increments to weights using~\Cref{eq:adv}, then fine-tune $\pi_0$ with
\begin{equation}
    \mathcal{L}_{\text{RWBC}}
    =
    \mathbb{E}_{p(a|o),q(a_t|a)}
    \left[
    \Delta_t \cdot
    \Norm{v_\theta(a_t, t \mid o) - (a - \epsilon)}^2
    \right],
    \label{eq:rwbc_loss}
\end{equation}
where $a_t = (1-t)\epsilon + ta$, $t \sim \mathcal{U}(0,1)$, $\epsilon \sim \mathcal{N}(0,I)$, and $\Delta_t$ is the \rwname-derived positive weight computed with $\beta = 0.2$ and $\delta_{\max} = 2.0$ in~\Cref{eq:adv}. We compare TOP-RWBC against behavior cloning (BC) on the same data, measuring partial success as the fraction of predefined subtasks completed per trial, summed over 10 trials. As shown in~\Cref{tab:real_world_awr}, TOP-RWBC improves over BC on all six tasks.

\begin{table}[t]
\centering
\footnotesize
\setlength{\tabcolsep}{4pt}
\caption{\textbf{Real-world experiments.} Partial success score out of 10 trials for reward-weighted behavior cloning on single-arm SO-100 tasks.}
\begin{tabular}{lccc}
\toprule
\textbf{Task} & \textbf{Pretrained} & \textbf{BC} & \textbf{TOP-RWBC (Ours)} \\
\midrule
Place toy car in box & 1 & 2 & \textbf{3} \\
Stack red cube on green cube & 1.33 & 1 & \textbf{2.33} \\
Put pen into cup & 1.67 & 5.67 & \textbf{6.33} \\
Place doll in box & 0 & 7 & \textbf{10} \\
Pick up cube & 4 & 7 & \textbf{10} \\
Put cube in cup & 4 & 6 & \textbf{9} \\
\bottomrule
\end{tabular}
\label{tab:real_world_awr}
\end{table}

\section{Limitations}
\label{sec:limitations}

\rwname\ inherits the visual perception limitations of the underlying VLM: tasks requiring fine-grained spatial reasoning, precise alignment, or small-object manipulation may receive noisy progress estimates when the model cannot visually distinguish intermediate states. The raw prefix score in~\Cref{eq:instr_reward} is causal, and the per-episode min-max normalization in~\Cref{eq:minmax} is non-causal and is used for within-trajectory evaluation and visualization. Finally, \rwname\ is sensitive to chat template formatting and answer-token choice, so deployments should validate the prompt and tokenizer behavior for the chosen VLM backbone.

\section{Conclusion}
\label{sec:conclusion}

We presented \rwname, a zero-shot progress reward method that repurposes token probabilities of pretrained video VLMs as instruction-conditioned progress signals for robotic manipulation. By querying the model's internal belief about instruction completion rather than requiring it to generate calibrated numerical outputs, \rwname\ sidesteps the well-known limitations of VLMs in numerical reasoning and instruction following. Across Open X-Embodiment and our newly introduced \dname\ benchmark, \rwname\ substantially outperforms GVL on open-source models. It also remains competitive with robotics reward models trained on large-scale data across progress-estimation metrics, while requiring no reward-model training. The instruction-disambiguation matrix and wrong-instruction raw-reward drop further show that rewards depend on the language instruction rather than prefix position alone. Finally, \rwname supports success detection without reward-model training, and \rwname\ can be used for reward-weighted behavior cloning, yielding consistent improvements over standard BC across six real-world SO-100 manipulation tasks.

\clearpage

\bibliography{main}

\newpage
\appendix
\onecolumn

% You can have as much text here as you want. The main body must be at most $8$
% pages long. For the final version, one more page can be added. If you want, you
% can use an appendix like this one.

% The $\mathtt{\backslash onecolumn}$ command above can be kept in place if you
% prefer a one-column appendix, or can be removed if you prefer a two-column
% appendix.  Apart from this possible change, the style (font size, spacing,
% margins, page numbering, etc.) should be kept the same as the main body.
%%%%%%%%%%%%%%%%%%%%%%%%%%%%%%%%%%%%%%%%%%%%%%%%%%%%%%%%%%%%%%%%%%%%%%%%%%%%%%%
%%%%%%%%%%%%%%%%%%%%%%%%%%%%%%%%%%%%%%%%%%%%%%%%%%%%%%%%%%%%%%

\section{Alternative Reward Formulation}
\label{appendix:alternative_reward}
In addition to the main formulation of \texttt{\rwname} presented in~\Cref{sec:method}, we also experimented with an alternative reward formulation that evaluates the probability of generating the entire instruction given the video trajectory. Specifically, we construct the following prompt,
\begin{quote}
`` \texttt{<|video|>} The above video shows a robot manipulation trajectory that completes the following task: \{instruction\}.''
\end{quote}

We then define the reward for a video prefix $\tau_{1:t}$ to be
\begin{equation}\label{eq:alt_instr_reward}
r_t \;=\; \sum_{i}\log p_\theta\!\left(\texttt{inst}_i \mid c(\tau_{1:t}, u, \texttt{inst}_{<i})\right),
\end{equation}
where $\texttt{inst}_i$ is the $i$-th token of the instruction, and $u$ represents the prompt between the video and the instruction. However, we found this alternative formulation to be less effective than the main formulation presented in~\Cref{sec:method}. We hypothesize that this is because the model will assign high probability to entities in the instruction if they ever appear in the robot video trajectory. For example, if there is an apple in the video, and the instruction is to peel the apple, then the model will assign high probability to \texttt{apple} in the instruction, defeating the purpose of progress estimation. In contrast, our main formulation only requires the model to judge whether the trajectory completes the instruction, which prevents such distraction in probability evaluation.

\section{Why the \texttt{True} Token?}
\label{appendix:true_token}
\begin{figure}[!htbp]
    \centering
    \includegraphics[width=0.95\linewidth]{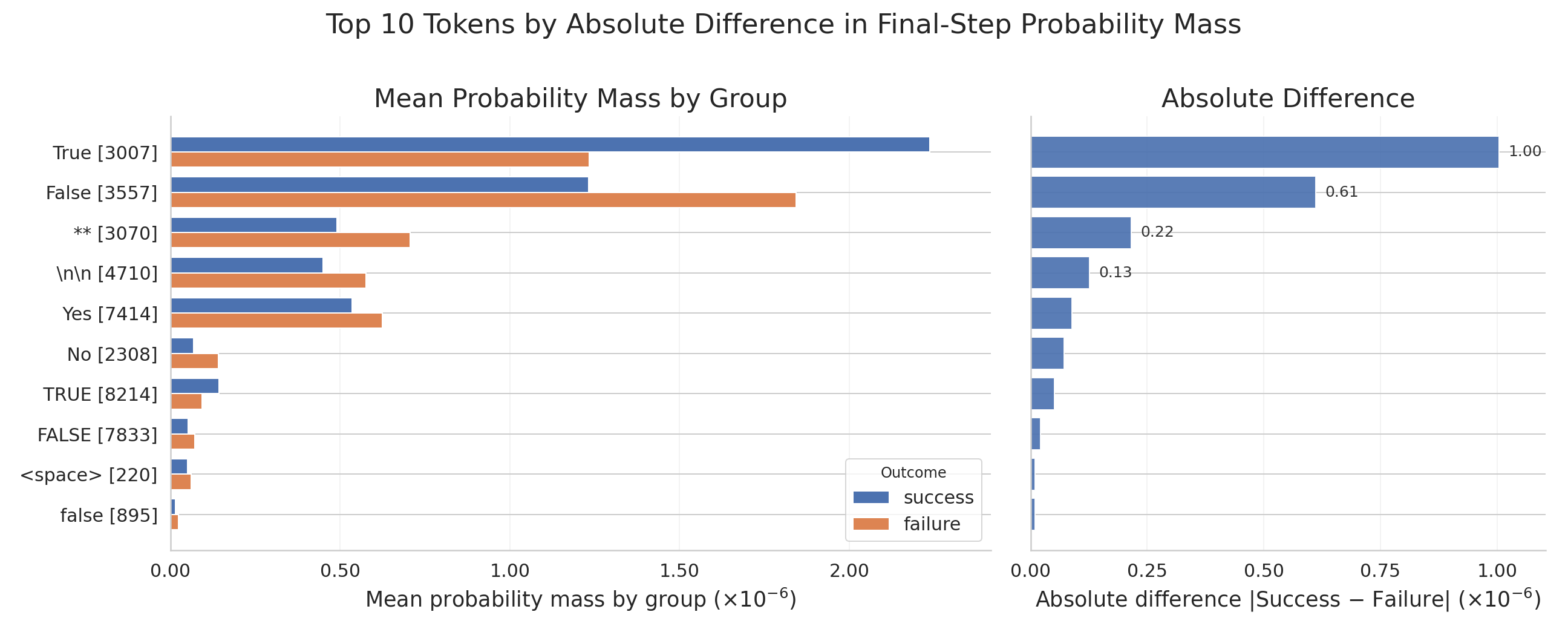}
    \caption{Top 10 tokens by absolute difference in mean final-step token probability between successful and failed trajectories. The affirmative token \texttt{True} shows the largest separation, motivating its use as the completion token in \rwname. Left: mean token probability by group; right: absolute difference in mean token probability.}
    \label{fig:appendix_true_token}
\end{figure}

We choose \texttt{True} as the affirmative completion token rather than alternatives (e.g., \texttt{Yes}) because it is a single token in our evaluated vocabularies and yields the largest, most consistent separation between successful and failed trajectories at the final step. \Cref{fig:appendix_true_token} shows the top tokens by absolute difference in mean final-step token probability; \texttt{True} exhibits the largest gap.

\section{Success Detection Score}
\label{app:success_score}
For success detection, we convert \rwname\ token probabilities into trajectory-level completion scores. Let $r_{t_k}^{+}=\log p_\theta(\texttt{True} \mid c(\tau_{1:t_k},x))$ and $r_{t_k}^{-}=\log p_\theta(\texttt{False} \mid c(\tau_{1:t_k},x))$ be the completion and non-completion log-probabilities for the $k$-th sampled prefix, and define the prefix completion margin $m_{t_k}=r_{t_k}^{+}-r_{t_k}^{-}$. We first measure how much this margin increases from the beginning to the end of the trajectory,
\begin{equation}
    s_{\text{top}}(\tau,x) =
    \frac{1}{|\mathcal{L}|}\sum_{k\in \mathcal{L}} m_{t_k}
    -
    \frac{1}{|\mathcal{E}|}\sum_{k\in \mathcal{E}} m_{t_k},
    \label{eq:success_topreward_score}
\end{equation}
where $\mathcal{E}$ and $\mathcal{L}$ denote the first and last three sampled prefixes, respectively. We also score the full video with a direct binary completion query,
\begin{equation}
    s_{\text{yn}}(\tau,x) =
    \log p_\theta(\texttt{Yes} \mid c_{\text{yn}}(\tau,x))
    -
    \log p_\theta(\texttt{No} \mid c_{\text{yn}}(\tau,x)),
    \label{eq:success_yesno_score}
\end{equation}
where $c_{\text{yn}}$ is the full-video prompt asking whether the trajectory completed the instruction. For both Qwen3-VL backbones, we use a fixed equal-weight combination of the standardized scores,
\begin{equation}
    s_{\text{succ}}(\tau,x) = z(s_{\text{top}}(\tau,x)) + z(s_{\text{yn}}(\tau,x)),
    \label{eq:success_combined_score}
\end{equation}
where $z(q)=(q-\mu_q)/\sigma_q$ is computed over the evaluated trajectories to put the two scores on comparable scale. This uses no learned classifier or fitted task-specific reward model.

\section{Dataset-level breakdown}
\label{app:dataset_breakdown}
\begin{figure}[!htbp]
    \centering
    \includegraphics[width=.55\linewidth]{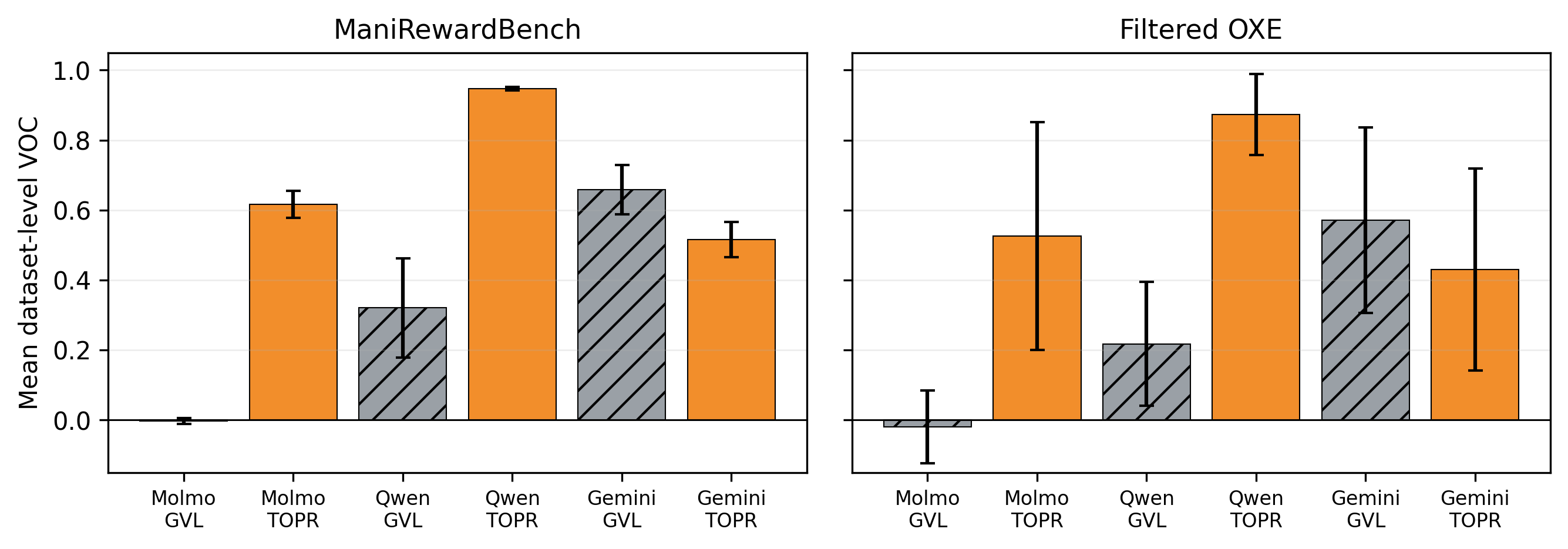}
    \caption{\textbf{VOC comparison across datasets.} Mean dataset-level VOC for GVL (0-shot) and \rwname\ across two evaluation sets: filtered OXE (33 datasets, 20 episodes each) and the \dname\ successful progress-evaluation subset (4 datasets, 113 tasks, 497 episodes). Error bars denote standard deviation across datasets within each evaluation set.}
    \label{fig:voc_comparison_bar}
\end{figure}

This section provides dataset-level details that complement the aggregate results in~\Cref{tab:oxe_voc,tab:bench_voc}. \Cref{fig:voc_comparison_bar} summarizes dataset-level VOC, and \Cref{tab:dataset_comparison_by_model} reports dataset-level VOC for GVL (0-shot) and $\rwname$ (TOPR) for each dataset and model backbone, using the same 33 high-quality OXE datasets as the main table.

\begin{table}[!ht]
    \centering
    \small
    \setlength{\tabcolsep}{2pt}
    \resizebox{\textwidth}{!}{\begin{tabular}{lrrrrrrrrr}
\toprule
\textbf{Dataset} & \multicolumn{3}{c}{\textbf{Qwen3-VL-8B}} & \multicolumn{3}{c}{\textbf{Gemini-2.5-Pro}} & \multicolumn{3}{c}{\textbf{Molmo2-8B}}\\
 & GVL & TOPR & $\Delta$ & GVL & TOPR & $\Delta$ & GVL & TOPR & $\Delta$\\
\midrule
\multicolumn{10}{l}{\textbf{ManiRewardBench}}\\
\texttt{ManiRewardBench\_bimanual\_yam} & 0.164 & \textbf{0.947} & 0.783 & \textbf{0.566} & 0.546 & -0.021 & 0.007 & \textbf{0.565} & 0.558\\
\texttt{ManiRewardBench\_franka} & 0.242 & \textbf{0.942} & 0.700 & \textbf{0.695} & 0.448 & -0.247 & 0.000 & \textbf{0.662} & 0.662\\
\texttt{ManiRewardBench\_lerobot} & 0.332 & \textbf{0.954} & 0.622 & \textbf{0.620} & 0.578 & -0.041 & -0.001 & \textbf{0.595} & 0.597\\
\texttt{ManiRewardBench\_single\_yam} & 0.544 & \textbf{0.945} & 0.401 & \textbf{0.752} & 0.488 & -0.264 & -0.017 & \textbf{0.642} & 0.659\\
\midrule
\multicolumn{10}{l}{\textbf{Open X-Embodiment}}\\
\texttt{aloha\_mobile\_cabinet} & 0.199 & \textbf{0.822} & 0.623 & \textbf{0.814} & 0.422 & -0.392 & 0.000 & \textbf{0.211} & 0.211\\
\texttt{aloha\_mobile\_wash\_pan} & 0.127 & \textbf{0.977} & 0.850 & \textbf{0.856} & 0.734 & -0.121 & 0.012 & \textbf{0.893} & 0.880\\
\texttt{aloha\_mobile\_wipe\_wine} & 0.243 & \textbf{0.961} & 0.718 & 0.699 & \textbf{0.825} & 0.125 & 0.000 & \textbf{0.804} & 0.804\\
\texttt{aloha\_static\_candy} & 0.414 & \textbf{0.963} & 0.549 & 0.697 & \textbf{0.803} & 0.105 & 0.000 & \textbf{0.842} & 0.842\\
\texttt{aloha\_static\_coffee} & 0.101 & \textbf{0.977} & 0.876 & \textbf{0.877} & 0.114 & -0.764 & 0.012 & \textbf{0.400} & 0.388\\
\texttt{aloha\_static\_cups\_open} & 0.271 & \textbf{0.837} & 0.565 & 0.524 & \textbf{0.765} & 0.241 & 0.000 & \textbf{0.864} & 0.864\\
\texttt{aloha\_static\_pro\_pencil} & 0.000 & \textbf{0.963} & 0.963 & \textbf{0.457} & -0.029 & -0.486 & 0.000 & \textbf{0.225} & 0.225\\
\texttt{aloha\_static\_screw\_driver} & 0.212 & \textbf{0.937} & 0.725 & \textbf{0.833} & 0.171 & -0.662 & 0.000 & \textbf{0.925} & 0.925\\
\texttt{aloha\_static\_vinh\_cup} & 0.332 & \textbf{0.916} & 0.584 & \textbf{0.836} & 0.723 & -0.113 & 0.000 & \textbf{0.971} & 0.971\\
\texttt{aloha\_static\_vinh\_cup\_left} & -0.057 & \textbf{0.903} & 0.960 & 0.458 & \textbf{0.898} & 0.441 & -0.050 & \textbf{0.707} & 0.757\\
\texttt{aloha\_static\_ziploc\_slide} & 0.301 & \textbf{0.978} & 0.677 & \textbf{0.845} & 0.394 & -0.452 & 0.000 & \textbf{0.918} & 0.918\\
\texttt{austin\_buds\_dataset} & 0.347 & \textbf{0.954} & 0.607 & \textbf{0.777} & -0.055 & -0.832 & 0.000 & \textbf{0.908} & 0.908\\
\texttt{austin\_sirius\_dataset} & 0.242 & \textbf{0.854} & 0.612 & \textbf{0.704} & 0.701 & -0.003 & 0.000 & \textbf{0.119} & 0.119\\
\texttt{berkeley\_fanuc\_manipulation} & 0.201 & \textbf{0.866} & 0.665 & \textbf{0.422} & 0.333 & -0.089 & \textbf{0.000} & -0.077 & -0.077\\
\texttt{berkeley\_rpt} & 0.399 & \textbf{0.983} & 0.584 & 0.303 & \textbf{0.600} & 0.297 & 0.000 & \textbf{0.605} & 0.605\\
\texttt{berkeleymvp} & 0.240 & \textbf{0.966} & 0.727 & \textbf{0.744} & 0.564 & -0.180 & 0.000 & \textbf{0.472} & 0.472\\
\texttt{cmu\_franka\_exploration\_dataset} & 0.000 & \textbf{0.626} & 0.626 & \textbf{0.622} & 0.272 & -0.350 & 0.000 & \textbf{0.325} & 0.325\\
\texttt{dlr\_edan\_shared\_control} & 0.166 & \textbf{0.867} & 0.700 & \textbf{0.651} & 0.583 & -0.068 & -0.600 & \textbf{0.770} & 1.370\\
\texttt{dlr\_sara\_grid\_clamp} & 0.000 & \textbf{0.740} & 0.740 & \textbf{0.253} & -0.166 & -0.419 & 0.000 & \textbf{0.562} & 0.562\\
\texttt{jaco\_play} & 0.090 & \textbf{0.907} & 0.818 & \textbf{0.513} & 0.508 & -0.005 & \textbf{0.000} & -0.299 & -0.299\\
\texttt{kaist\_nonprehensile} & 0.158 & \textbf{0.914} & 0.756 & \textbf{0.491} & 0.349 & -0.142 & 0.000 & \textbf{0.251} & 0.251\\
\texttt{nyudoor} & 0.608 & \textbf{0.802} & 0.194 & \textbf{0.872} & 0.729 & -0.142 & 0.000 & \textbf{0.650} & 0.650\\
\texttt{nyufranka} & 0.232 & \textbf{0.875} & 0.642 & \textbf{0.772} & 0.494 & -0.278 & 0.059 & \textbf{0.865} & 0.806\\
\texttt{stanford\_hydra\_dataset} & 0.164 & \textbf{0.973} & 0.809 & 0.379 & \textbf{0.557} & 0.178 & 0.000 & \textbf{0.091} & 0.091\\
\texttt{stanford\_kuka\_multimodal\_dataset} & 0.122 & \textbf{0.821} & 0.700 & -0.390 & \textbf{-0.055} & 0.335 & 0.000 & \textbf{0.493} & 0.493\\
\texttt{stanford\_robocook} & 0.035 & \textbf{0.443} & 0.408 & 0.329 & \textbf{0.521} & 0.191 & 0.000 & \textbf{0.582} & 0.582\\
\texttt{taco\_play} & 0.015 & \textbf{0.779} & 0.764 & \textbf{0.050} & 0.024 & -0.026 & 0.000 & \textbf{0.342} & 0.342\\
\texttt{tokyo\_u\_lsmo} & 0.215 & \textbf{0.977} & 0.762 & \textbf{0.690} & 0.501 & -0.188 & 0.000 & \textbf{0.719} & 0.719\\
\texttt{ucsd\_kitchen\_dataset} & 0.183 & \textbf{0.718} & 0.536 & \textbf{0.542} & -0.006 & -0.547 & 0.000 & \textbf{0.228} & 0.228\\
\texttt{ucsd\_pick\_and\_place\_dataset} & 0.090 & \textbf{0.819} & 0.729 & \textbf{0.683} & 0.520 & -0.163 & 0.000 & \textbf{0.605} & 0.605\\
\texttt{utokyo\_pr2\_opening\_fridge} & 0.485 & \textbf{0.957} & 0.472 & 0.372 & \textbf{0.644} & 0.272 & -0.075 & \textbf{0.849} & 0.924\\
\texttt{utokyo\_pr2\_tabletop\_manipulation} & 0.784 & \textbf{0.946} & 0.162 & \textbf{0.696} & 0.485 & -0.211 & 0.000 & \textbf{0.121} & 0.121\\
\texttt{utokyo\_xarm\_bimanual} & 0.266 & \textbf{0.815} & 0.549 & \textbf{0.495} & 0.277 & -0.218 & 0.000 & \textbf{0.386} & 0.386\\
\bottomrule
\end{tabular}
}
    \caption{Filtered datasets (alphabetical) comparing GVL (0-shot) vs $\rwname$ (TOPR) and their difference, broken down by model backbone. Bold indicates the higher VOC between the two methods for that dataset/model.}
    \label{tab:dataset_comparison_by_model}
\end{table}

\begin{figure}[!htbp]
    \centering
    \includegraphics[width=1\linewidth]{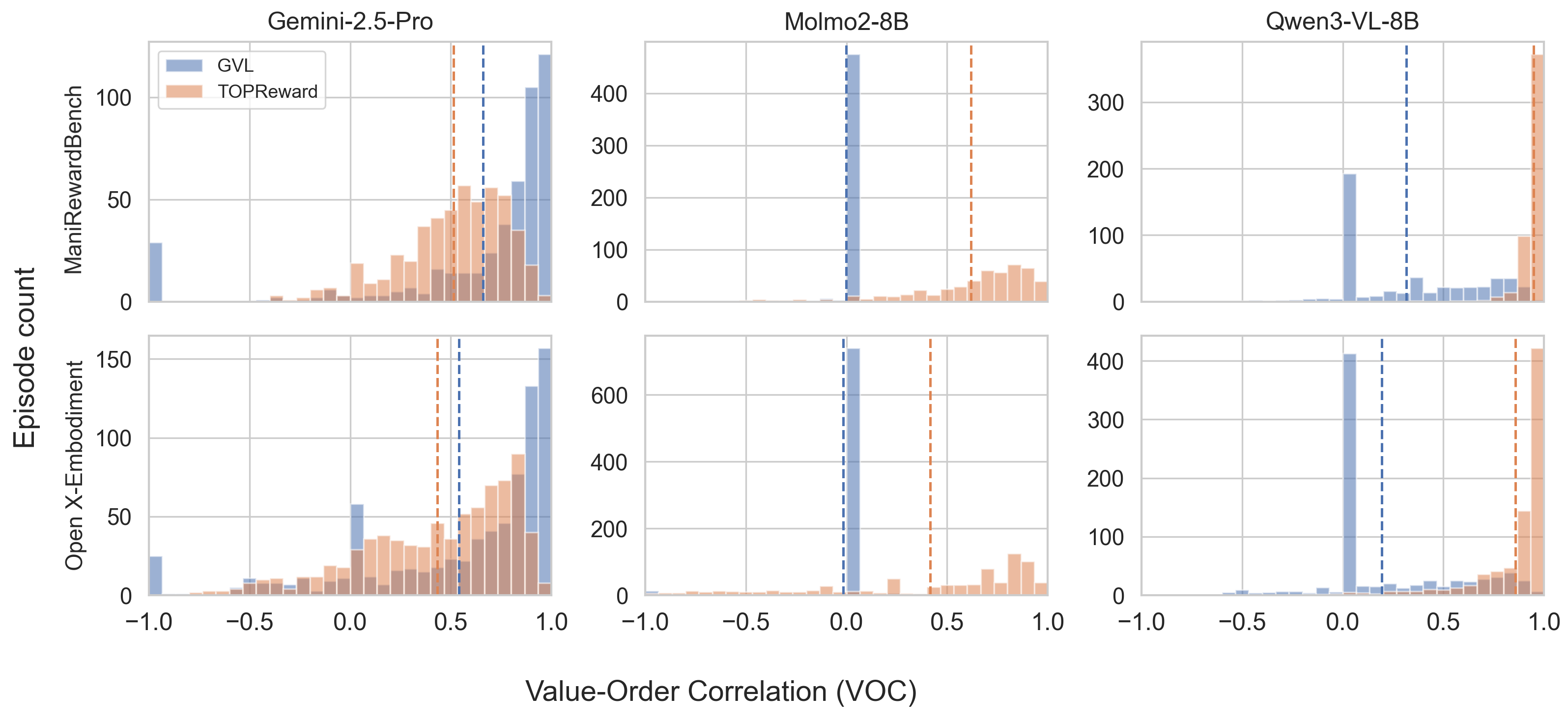}
    \caption{Per-episode VOC distributions, broken down by evaluation set (\dname\ vs Open X-Embodiment) and model backbone.}
    \label{fig:appendix_voc_overall_grid}
\end{figure}

\begin{figure}[!htbp]
    \centering
    \includegraphics[width=0.9\linewidth]{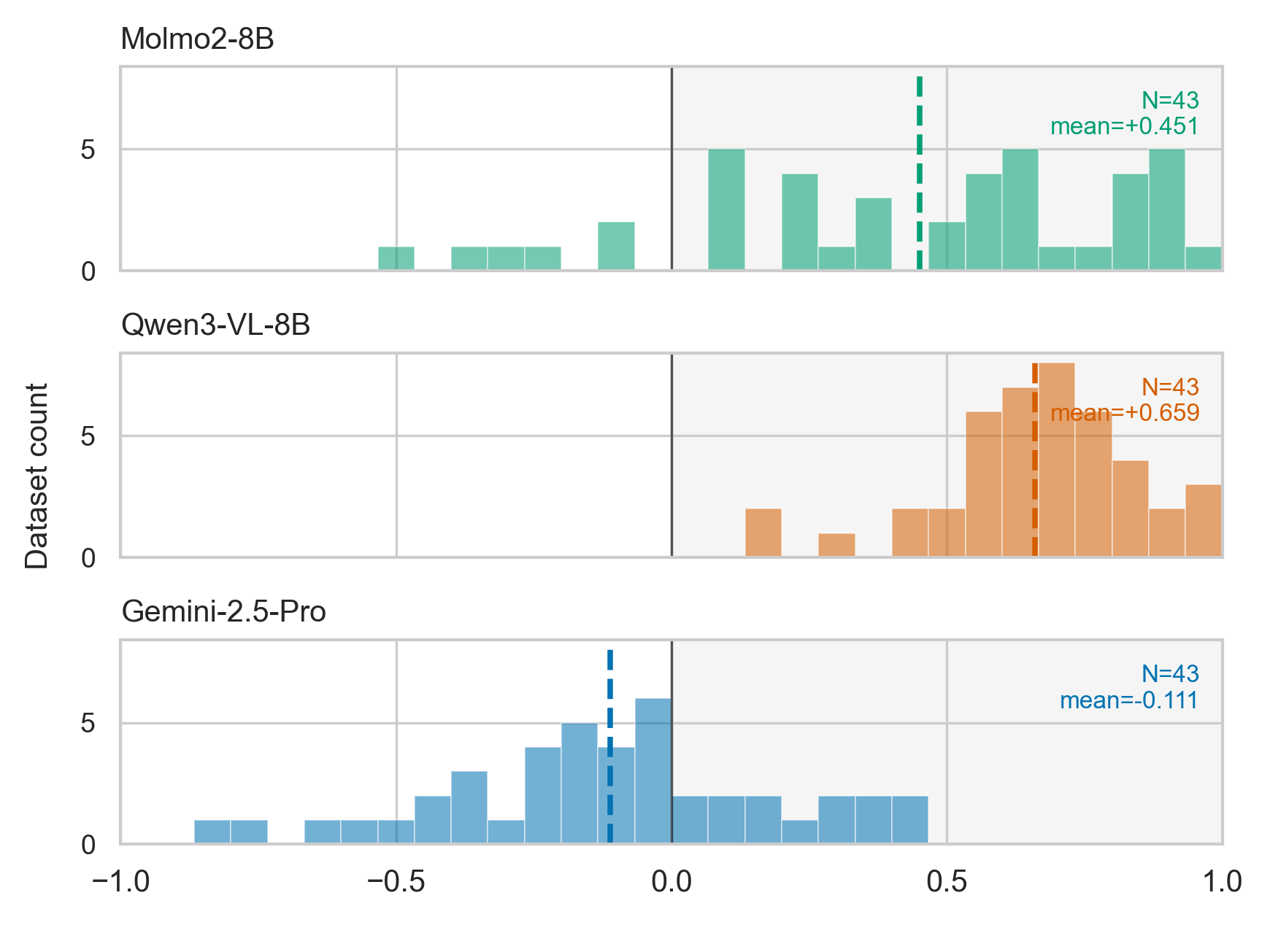}
    \caption{Distribution of dataset-level $\Delta$VOC $=$ VOC(\rwname) $-$ VOC(GVL), shown separately for each model backbone. Positive values indicate \rwname\ outperforms GVL; the dashed line marks the per-model mean.}
    \label{fig:appendix_delta_voc_histogram}
\end{figure}

\section{Additional qualitative results and ablations}
\label{app:additional_qualitative}

\begin{figure}[!htbp]
    \centering
    \includegraphics[width=\linewidth]{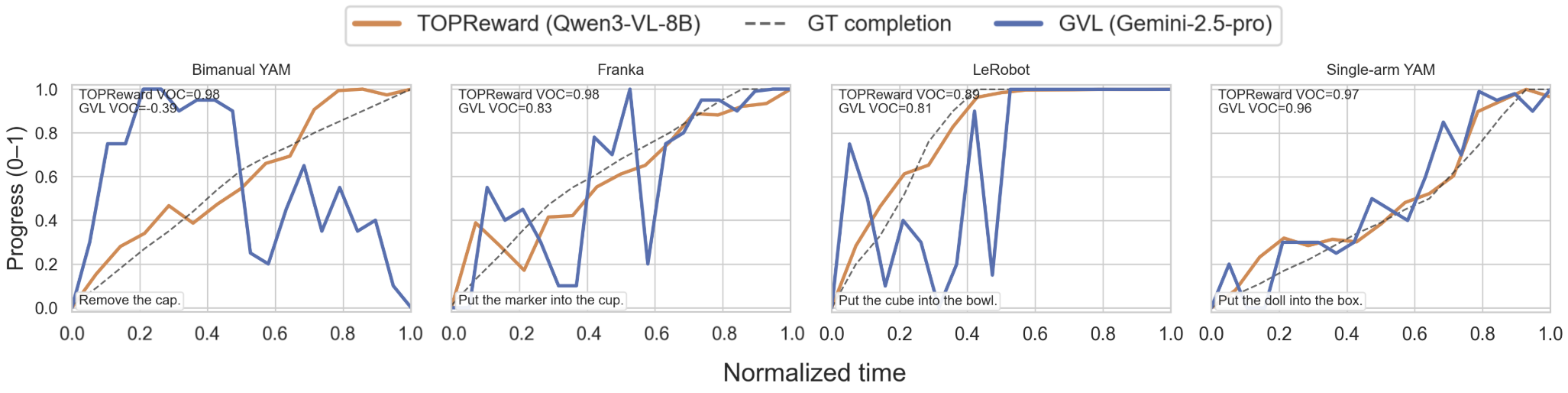}
    \caption{\textbf{Progress traces for ManiRewardBench.} Example progress traces predicted by \rwname\ (orange) compared to stage-aware ground-truth completion (dashed) from \dname, computed from annotated subtask boundaries. We also overlay Gemini-GVL (blue) on the same episodes when available.}
    \label{fig:linear_progress_examples}
\end{figure}

\begin{figure}[!htbp]
    \centering
    \includegraphics[width=.5\linewidth]{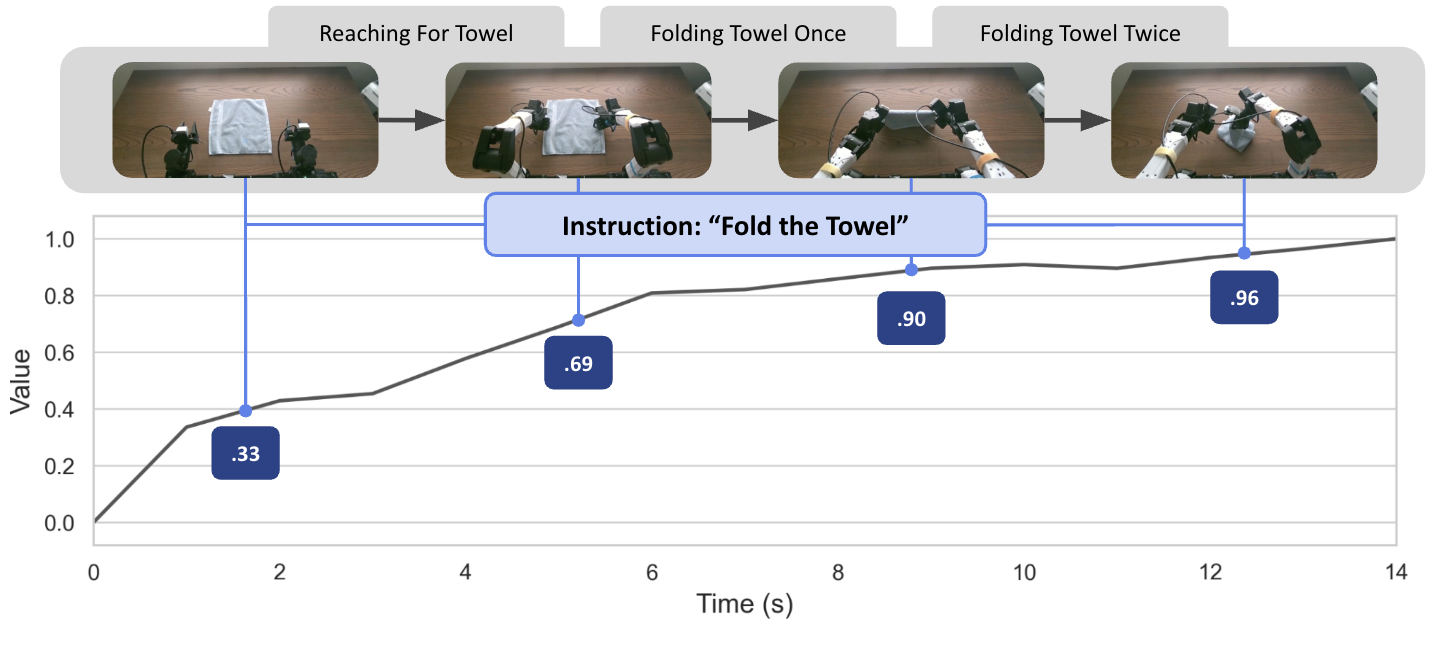}
    \caption{\textbf{Qualitative example of ``Fold the Towel'':} Instruction-conditioned progress estimation on a real trajectory. The curve shows \rwname's predicted completion value over time, with annotated values at selected frames corresponding to semantic subtasks.}
    \label{fig:method_example}
\end{figure}

\subsection{Real-world policy details}
\label{app:real_world_details}

For the reward-weighted behavior cloning objective in~\Cref{eq:rwbc_loss}, we use $\beta = 0.2$ and $\delta_{\max} = 2.0$ in~\Cref{eq:adv}.

\begin{figure}[!htbp]
    \centering
    \includegraphics[width=0.99\linewidth]{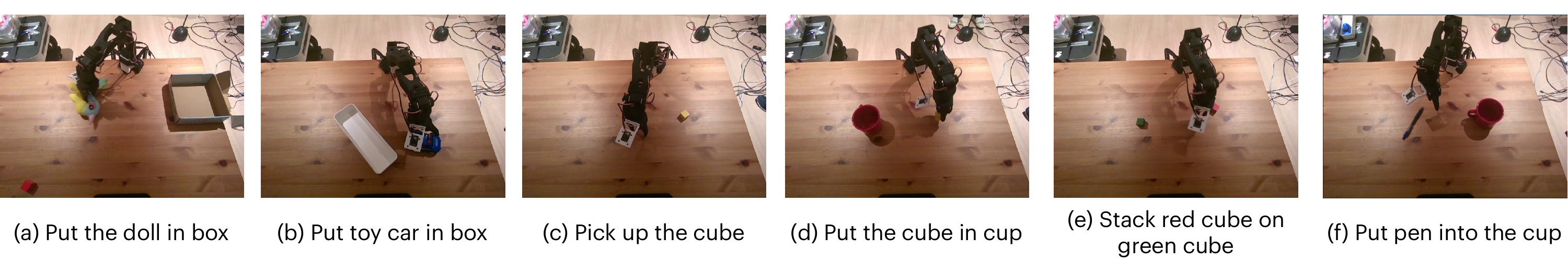}
    \caption{The six real-world single-arm SO-100 manipulation tasks used for reward-weighted behavior cloning evaluation.}
    \label{fig:real_world_exp}
\end{figure}

\begin{figure}[!htbp]
    \centering
    \includegraphics[width=\linewidth]{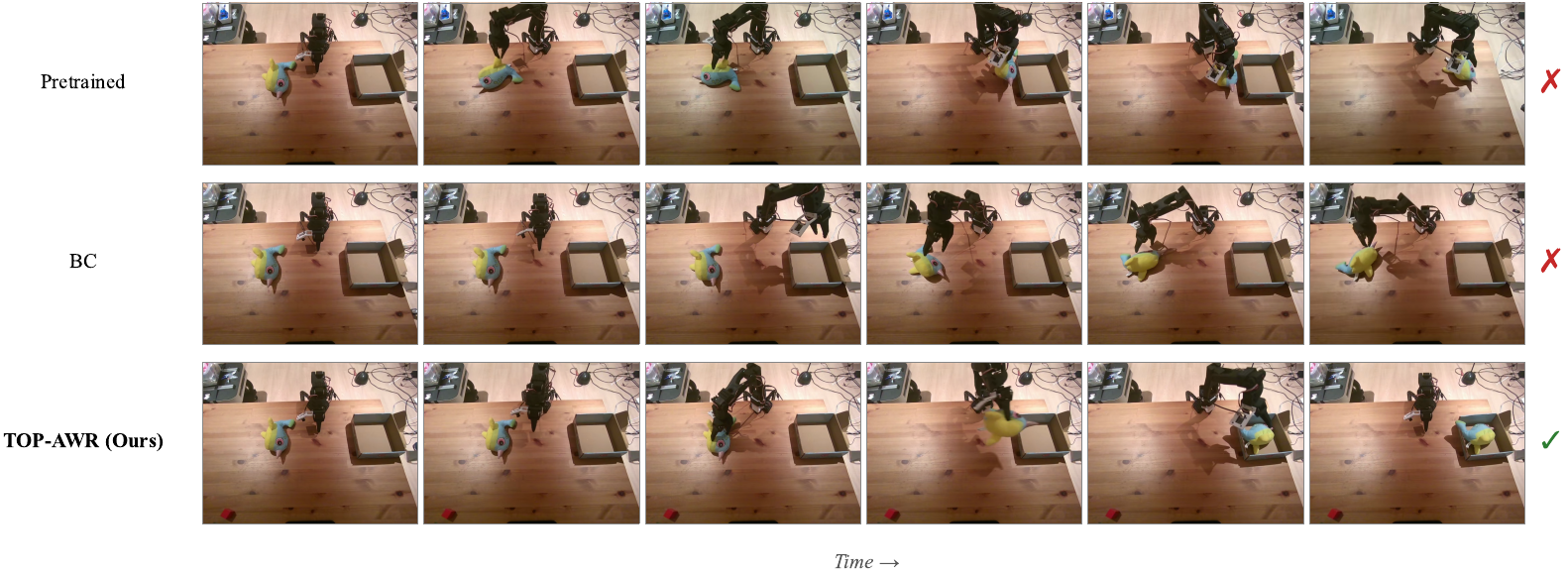}
    \caption{\textbf{Qualitative comparison on ``Place doll in box.''} The pretrained policy and behavior cloning (BC) both fail, while TOP-RWBC, fine-tuned with reward weights from \rwname, succeeds consistently. Frames are uniformly sampled from evaluation rollouts.}
    \label{fig:rollout_frame_strip}
\end{figure}

\subsection{Chat-template ablation}
\label{app:chat_ablation}
The Gemini API enforces a chat template, while our preferred open-source inference setting scores the prompt directly without one. To test this factor, we wrap the prompt in~\Cref{sec:probreward} with a chat template and evaluate the probability of the answer being \texttt{True}. \Cref{tab:chat_ablation} shows that chat formatting substantially reduces VOC for both Qwen3-VL-8B and Molmo2-8B.

\begin{table}[!htbp]
\centering
\small
\caption{\textbf{Effect of Chat Template on TOPReward VOC.} Chat template degrades Qwen3-VL-8B performance by nearly 50\% and Molmo2-8B by 20\%, demonstrating that the logit-based formulation is sensitive to prompt formatting.}
\label{tab:chat_ablation}
\begin{tabular}{lcccc}
\toprule
& \multicolumn{2}{c}{\textbf{Qwen3-VL-8B}} & \multicolumn{2}{c}{\textbf{Molmo2-8B}} \\
\cmidrule(lr){2-3} \cmidrule(lr){4-5}
\textbf{Dataset} & Base & +Chat & Base & +Chat \\
\midrule
Bimanual YAM       & 0.947 & 0.269 & 0.570 & 0.408 \\
Franka              & 0.943 & 0.528 & 0.696 & 0.615 \\
Single-arm YAM     & 0.946 & 0.703 & 0.691 & 0.546 \\
\midrule
\textbf{Mean}       & \textbf{0.945} & \textbf{0.500} & \textbf{0.652} & \textbf{0.523} \\
\textbf{$\Delta$}   & \multicolumn{2}{c}{\cellcolor{red!20}$-47.1\%$} & \multicolumn{2}{c}{\cellcolor{red!20}$-19.8\%$} \\
\bottomrule
\end{tabular}
\end{table}

\FloatBarrier
\section{Additional details of \dname} \label{app:add_bench}

The benchmark includes 130 diverse tasks that capture a wide range of everyday manipulation activities, such as stacking objects, sorting items, and interacting with containers. The dataset is manually collected using \textit{Franka}, \textit{SO-100/101}, \textit{bimanual YAM}, and \textit{single-arm YAM}.

Example challenging tasks in \dname include:

\paragraph{Multi-step / Reasoning tasks.}
\begin{itemize}
  \item \textit{``Push the puzzles to spell word GO''} (LeRobot) --- spatial reasoning combined with sequential multi-object manipulation.
  \item \textit{``Build a pyramid''} (Bimanual YAM) --- multi-step stacking with precise positioning, requiring four distinct subtasks.
  \item \textit{``Group the cubes by color''} (Bimanual YAM) --- requires color perception and categorical reorganization of multiple objects.
  \item \textit{``Put the cubes of the same colors together''} (Franka) --- color-conditional sorting across multiple objects.
  \item \textit{``Remove the block and stack the green cube on the red cube''} (LeRobot) --- obstacle removal followed by color-conditional stacking.
  \item \textit{``Pack and close the box''} (Franka) --- multi-phase task involving packing objects then closing the container.
\end{itemize}

\paragraph{Fine manipulation / Precise control.}
\begin{itemize}
  \item \textit{``Align the cubes horizontally''} (Bimanual YAM) --- fine spatial alignment, corresponding to the longest execution durations in the dataset.
  \item \textit{``Rotate the banana by 90 degrees''} / \textit{``Rotate the marker by 45 degrees''} (Franka) --- precise rotation control with specified angles.
  \item \textit{``Make the screw points to the glue''} (Bimanual YAM) --- precise orientation alignment of two distinct objects.
  \item \textit{``Pour tea''} (Franka) --- requires controlled pouring motion and spatial orientation awareness.
\end{itemize}

\paragraph{Deformable object handling.}
\begin{itemize}
  \item \textit{``Fold the towel''} / \textit{``Fold towel''} (Franka / Bimanual YAM) --- deformable material manipulation requiring careful grasp and fold planning.
  \item \textit{``Stack one cloth on top of another''} (Single-arm YAM) --- soft object stacking with non-rigid geometry.
\end{itemize}

\paragraph{Abstract / Symbolic tasks.}
\begin{itemize}
  \item \textit{``Press enter and then space key''} (Franka) --- keyboard interaction requiring sequential key presses.
  \item \textit{``Set table''} (Bimanual YAM) --- open-ended task requiring understanding of table-setting conventions.
\end{itemize}

The following table summarizes the statistics of each dataset in \dname.

\begin{table}[ht]
  \centering
  \caption{Summary of \dname\ datasets. 6 tasks appear in both the Lerobot and Lerobot failed splits, giving 130 unique tasks across the full benchmark.}
  \label{tab:dataset_summary}
  \begin{tabular}{lcc}
    \toprule
    \textbf{Dataset} & \textbf{Episodes} & \textbf{Tasks} \\
    \midrule
    Lerobot & 150 & 22 \\
    Lerobot failed & 156 & 23 \\
    Franka & 150 & 51 \\
    Bimanual YAM & 97 & 20 \\
    Single-arm YAM & 100 & 20 \\
    \midrule
    \textbf{Total} & \textbf{653} & \textbf{136} \\
    \bottomrule
  \end{tabular}
  \vspace{0.5em}
\end{table}

We briefly describe each dataset below:
\begin{itemize}
  \item \textbf{Lerobot Bimanual dataset}: Successful bimanual LeRobot manipulation demos (push, put, remove, stack tasks), 5--10 episodes per task.
  \item \textbf{Lerobot failure dataset}: Mixed failed and successful trajectory examples with the same task types, $\sim$7 episodes per task.
  \item \textbf{Franka dataset}: Franka robot demos with a diverse set of 51 instructions (rotation, cleaning, packing, pick-and-place), mostly 3 episodes per task.
  \item \textbf{Bimanual YAM dataset}: Bimanual YAM manipulation (fold, stack, build, open, etc.), 5 episodes per task.
  \item \textbf{Single-arm YAM dataset}: Single-arm YAM manipulation (put, remove, stack), 5 episodes per task.
\end{itemize}
% \begin{figure}[ht]
%   \vskip 0.2in
%   \begin{center}
%     \centerline{\includegraphics[width=\columnwidth]{number_of_episodes_per_task.png}}
%     \caption{
%       Number of episodes per example task by robot platform
%     }
%     \label{icml-historical}
%   \end{center}
% \end{figure}
\begin{figure}[ht]
  \vskip 0.2in
  \begin{center}
    \centerline{\includegraphics[width=\columnwidth]{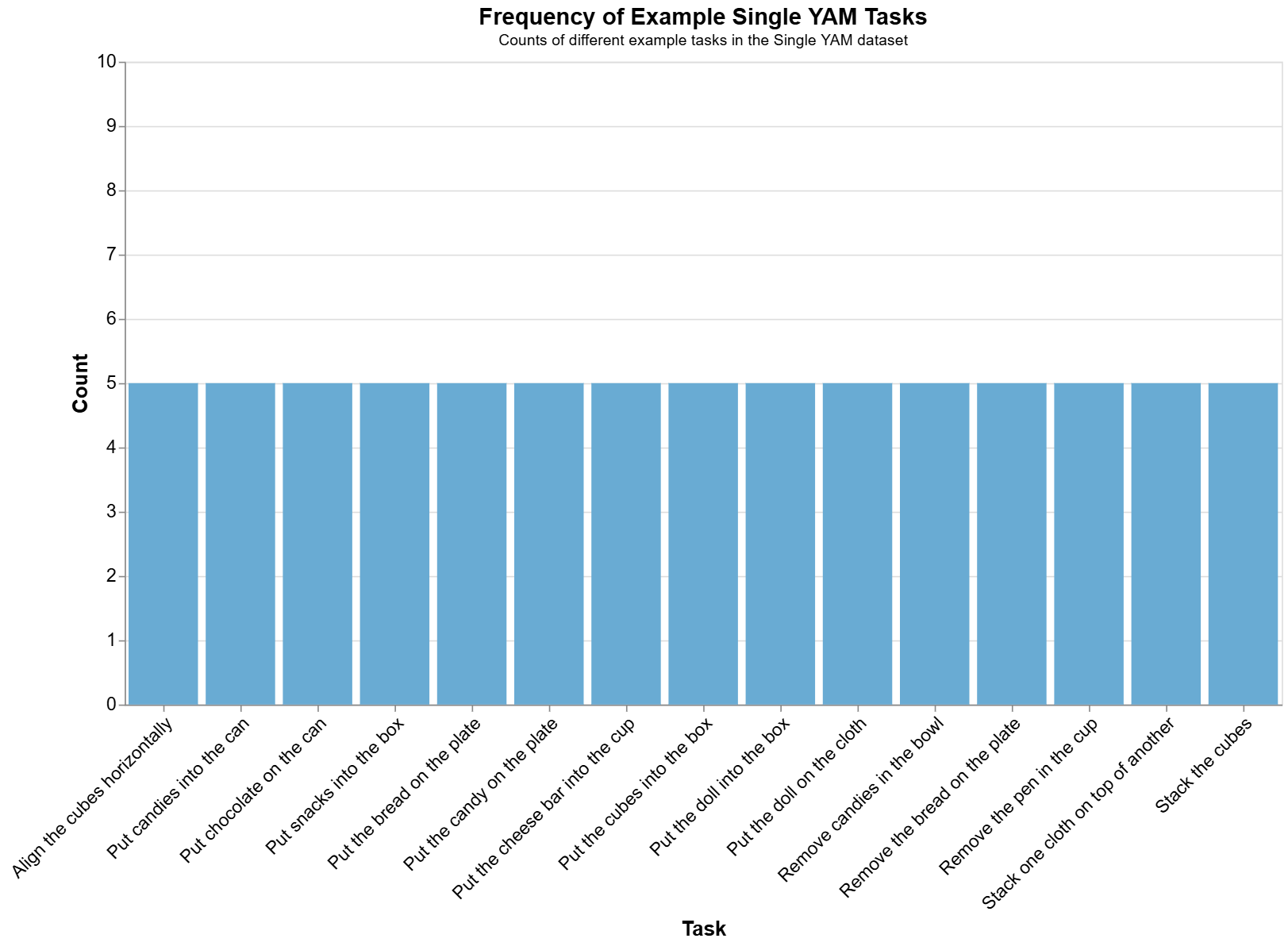}}
    \caption{
      Counts of different example tasks in the single-arm YAM dataset.
    }
    \label{fig:single_yam_tasks}
  \end{center}
\end{figure}

\begin{figure}[ht]
  \vskip 0.2in
  \begin{center}
    \centerline{\includegraphics[width=\columnwidth]{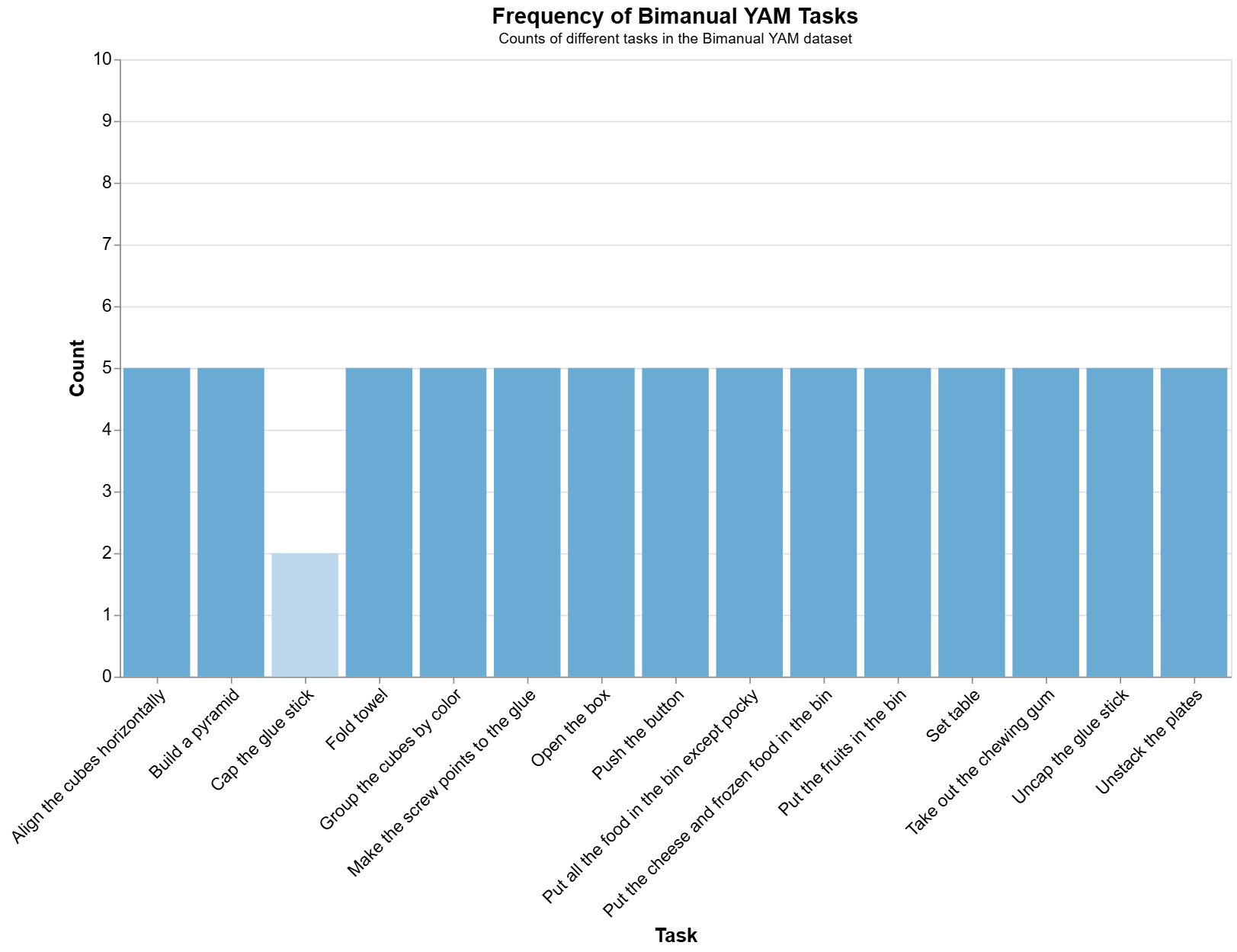}}
    \caption{
      Counts of different example tasks in the bimanual YAM dataset.
    }
    \label{fig:bimanual_yam_tasks}
  \end{center}
\end{figure}

\begin{figure}[ht]
  \vskip 0.2in
  \begin{center}
    \centerline{\includegraphics[width=\columnwidth]{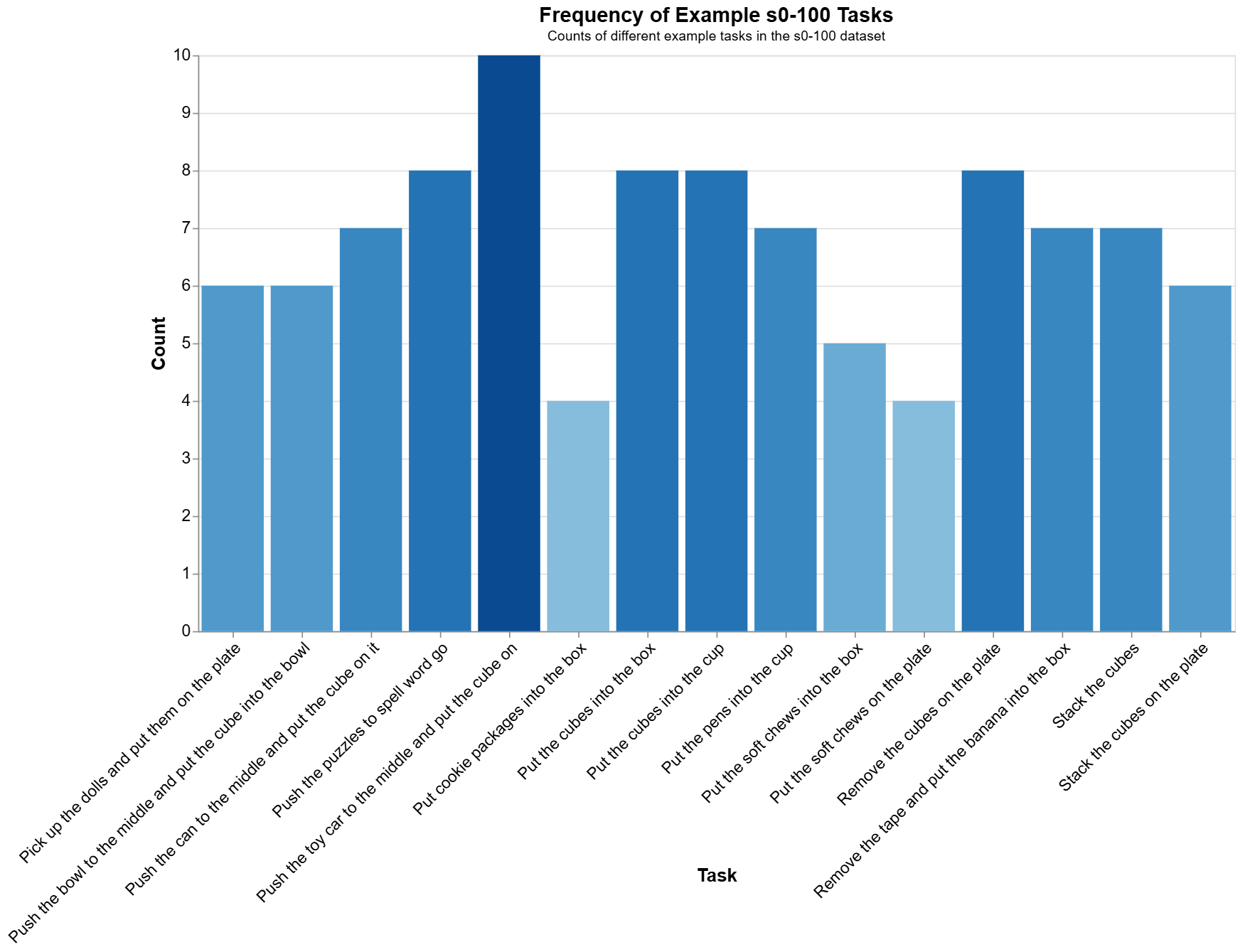}}
    \caption{
      Counts of different example tasks in the SO-100 dataset.
    }
    \label{fig:s0_100_tasks}
  \end{center}
\end{figure}

\begin{figure}[ht]
  \vskip 0.2in
  \begin{center}
    \centerline{\includegraphics[width=\columnwidth]{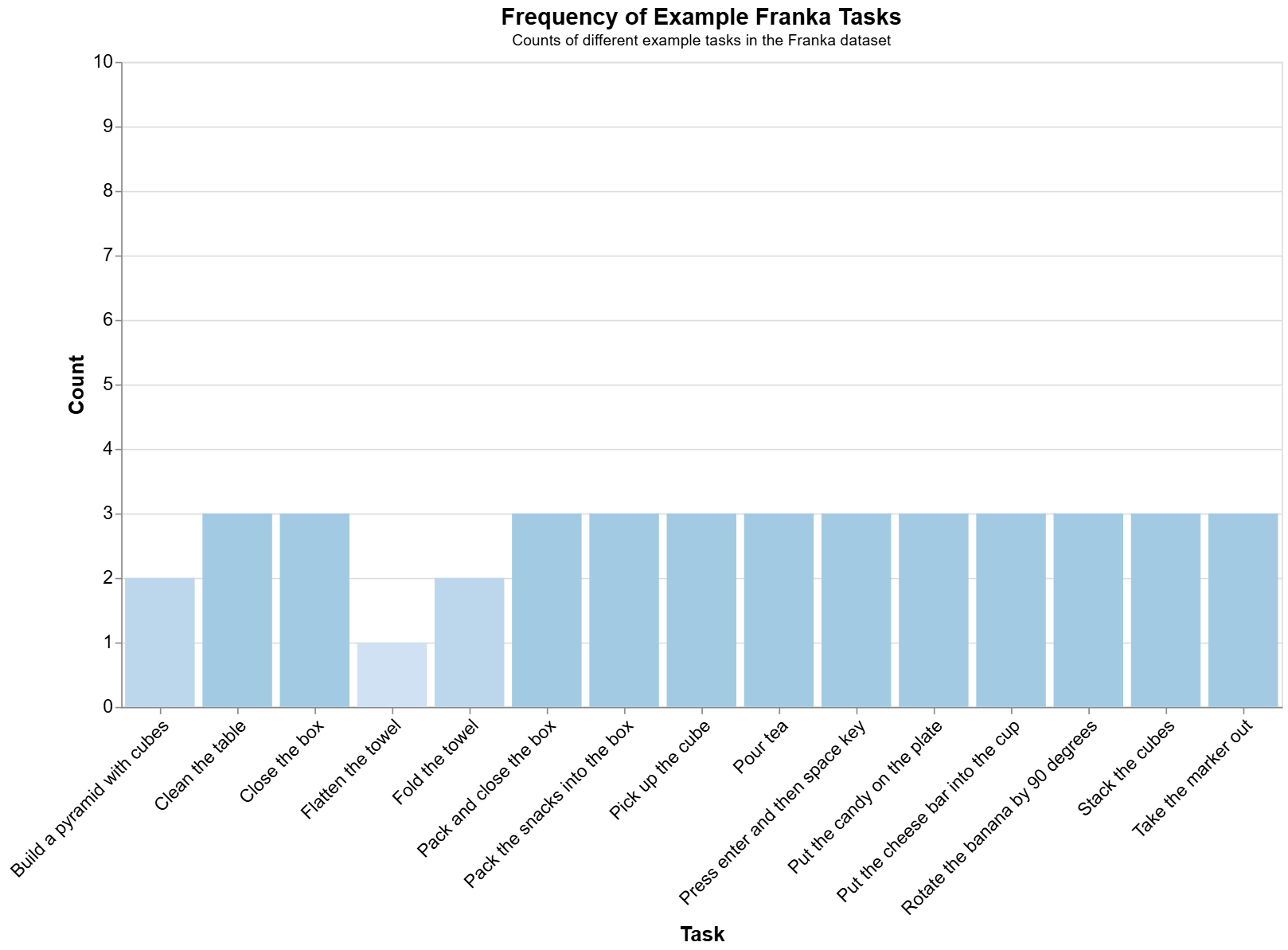}}
    \caption{
      Counts of different example tasks in the Franka dataset.
    }
    \label{fig:franka_tasks}
  \end{center}
\end{figure}

\begin{figure}[ht]
  \vskip 0.2in
  \begin{center}
    \centerline{\includegraphics[width=\columnwidth]{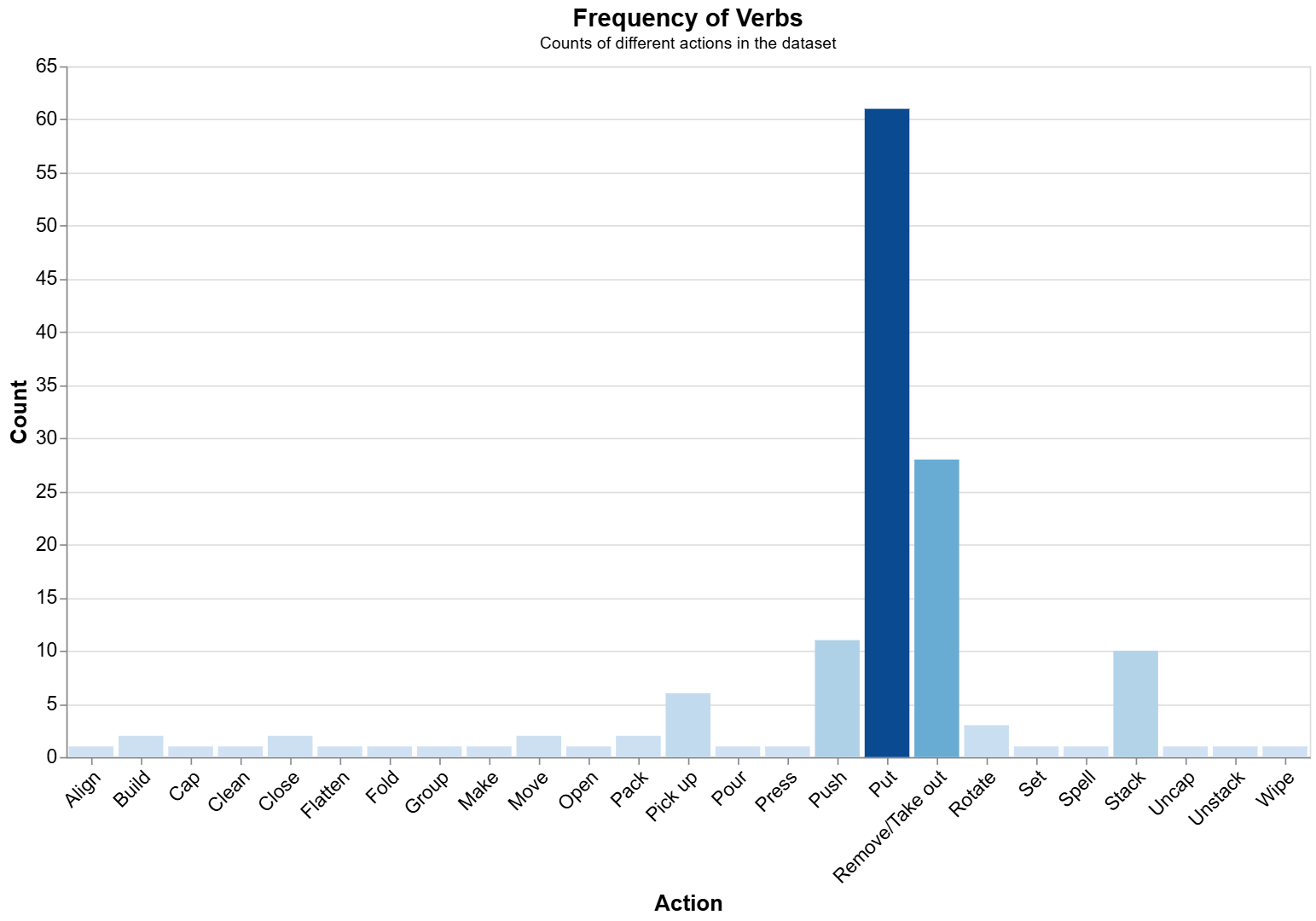}}
    \caption{
      Frequency of verbs
    }
    \label{fig:frequency_verbs}
  \end{center}
\end{figure}

\subsection{Subtask Annotation}
For each task, episodes are manually labeled and segmented into a sequence of predefined subtasks. Each task is associated with an ordered list of subtasks that represent stages of execution (e.g., reaching for an object, grasping it, or placing it). For every subtask, annotators specify a \texttt{start\_second} (the time in seconds when the subtask begins) and an \texttt{end\_second} (the time in seconds when it ends). Subtasks are non-overlapping and strictly ordered in time, with each subtask beginning immediately after the previous one ends.

\begin{figure}[ht]
  \vskip 0.2in
  \begin{center}
    % First row
    \begin{subfigure}[b]{0.45\columnwidth}
        \centering
        % Inner images using minipage, no subfigure
        \includegraphics[width=\linewidth]{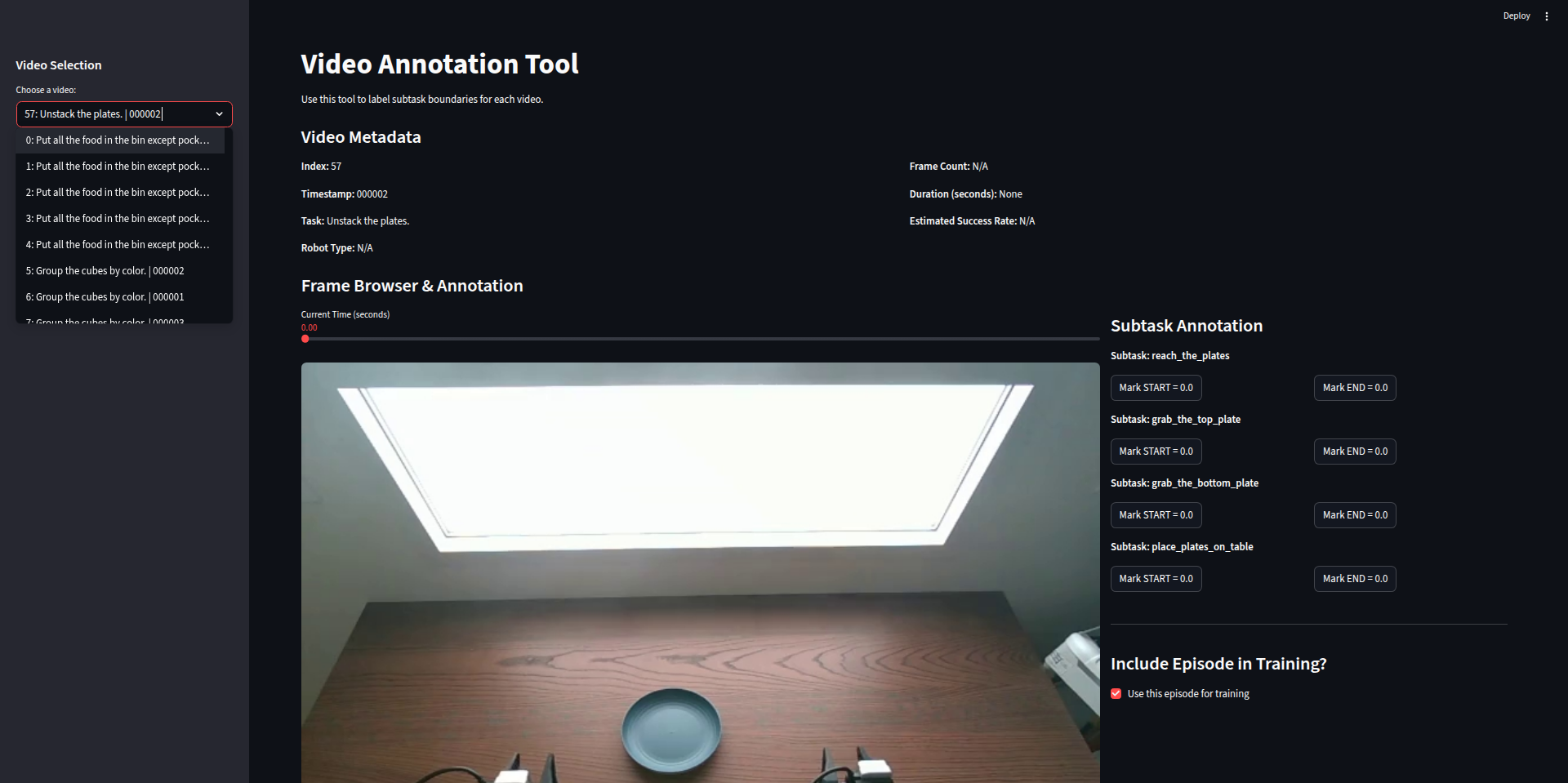}
    \end{subfigure}
    \begin{subfigure}[b]{0.45\columnwidth}
        \centering
        \includegraphics[width=\linewidth]{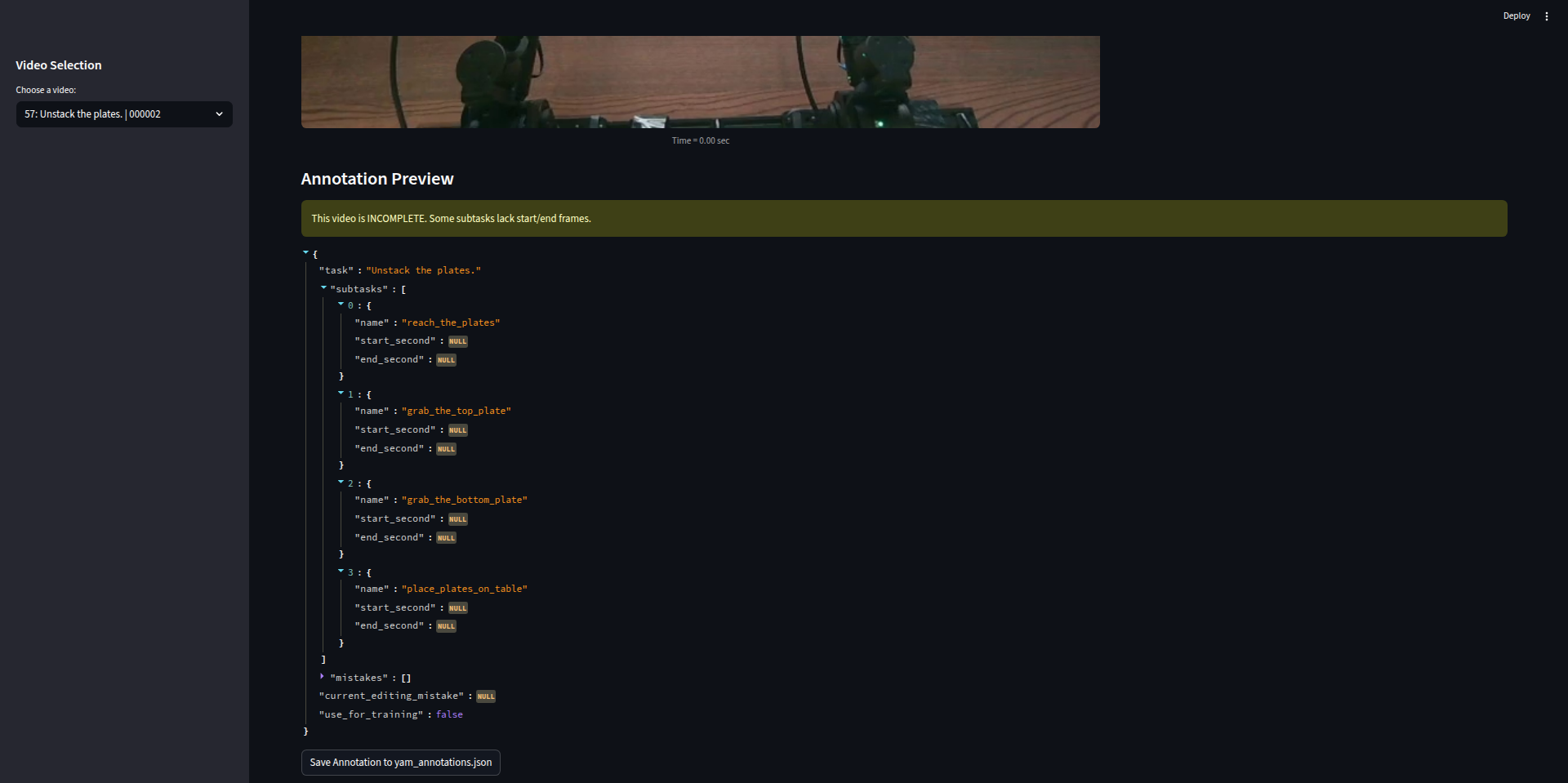}
    \end{subfigure}

    \caption{Screenshot of the Annotation Tool.}
    \label{fig:annotation_tool}
  \end{center}
\end{figure}

\begin{figure}[ht]
  \vskip 0.2in
  \begin{center}
    % First row
    \begin{subfigure}[b]{0.45\columnwidth}
        \centering
        % Inner images using minipage, no subfigure
        \begin{minipage}[b]{0.45\linewidth}
            \centering
            \includegraphics[width=\linewidth]{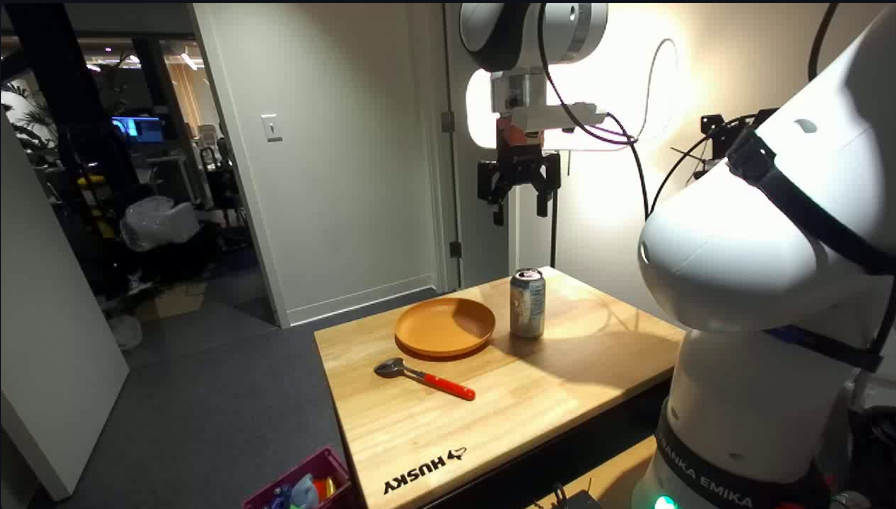}
            \caption*{\texttt{start\_second}: 0.0} % caption*, doesn't count
        \end{minipage}
        \begin{minipage}[b]{0.45\linewidth}
            \centering
            \includegraphics[width=\linewidth]{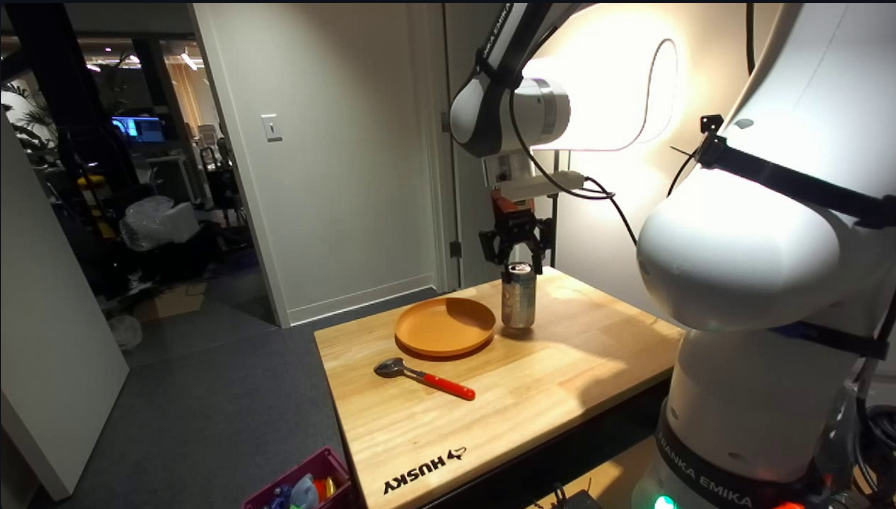}
            \caption*{\texttt{end\_second}: 3.9}
        \end{minipage}
        \caption{Grab the can}
        \label{fig:step1}
    \end{subfigure}
    \begin{subfigure}[b]{0.45\columnwidth}
        \centering
        \begin{minipage}[b]{0.45\linewidth}
            \centering
            \includegraphics[width=\linewidth]{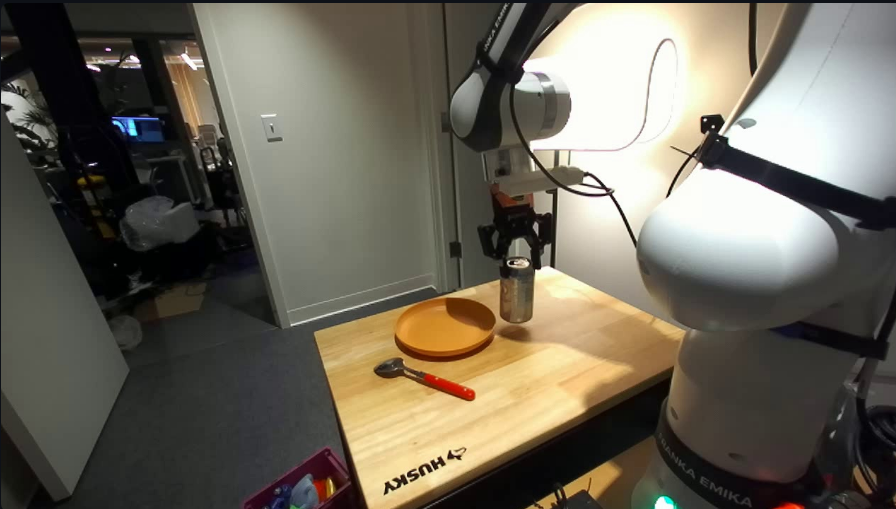}
            \caption*{\texttt{start\_second}: 4.0}
        \end{minipage}
        \begin{minipage}[b]{0.45\linewidth}
            \centering
            \includegraphics[width=\linewidth]{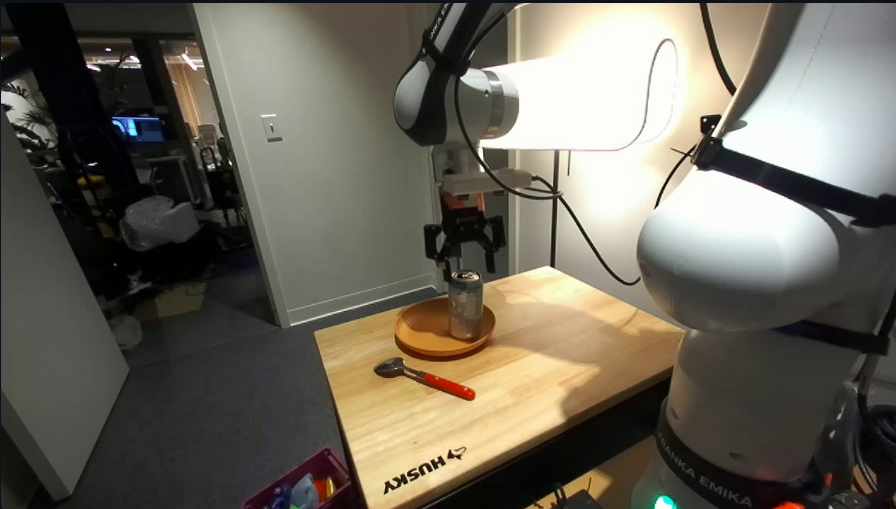}
            \caption*{\texttt{end\_second}: 6.4}
        \end{minipage}
        \caption{Place the can in the plate}
        \label{fig:step2}
    \end{subfigure}

    \vskip 0.3in

    % Second row
    \begin{subfigure}[b]{0.45\columnwidth}
        \centering
        \begin{minipage}[b]{0.45\linewidth}
            \centering
            \includegraphics[width=\linewidth]{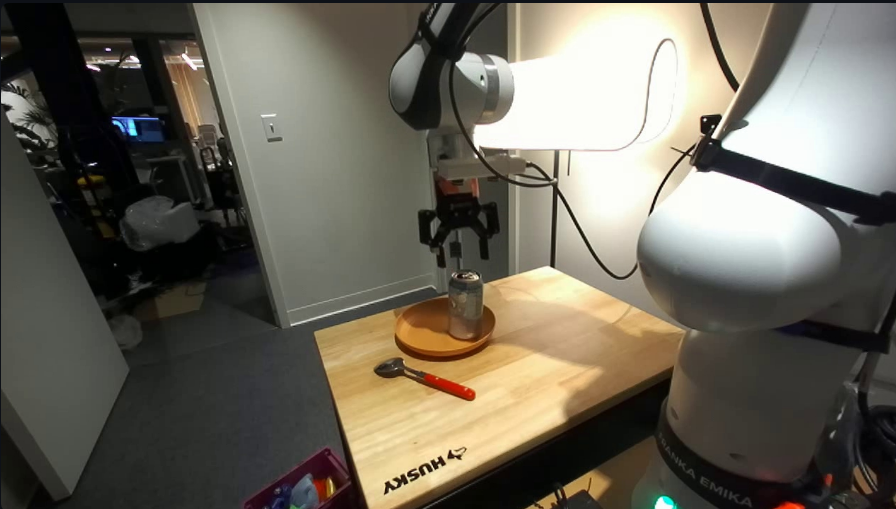}
            \caption*{\texttt{start\_second}: 6.5}
        \end{minipage}
        \begin{minipage}[b]{0.45\linewidth}
            \centering
            \includegraphics[width=\linewidth]{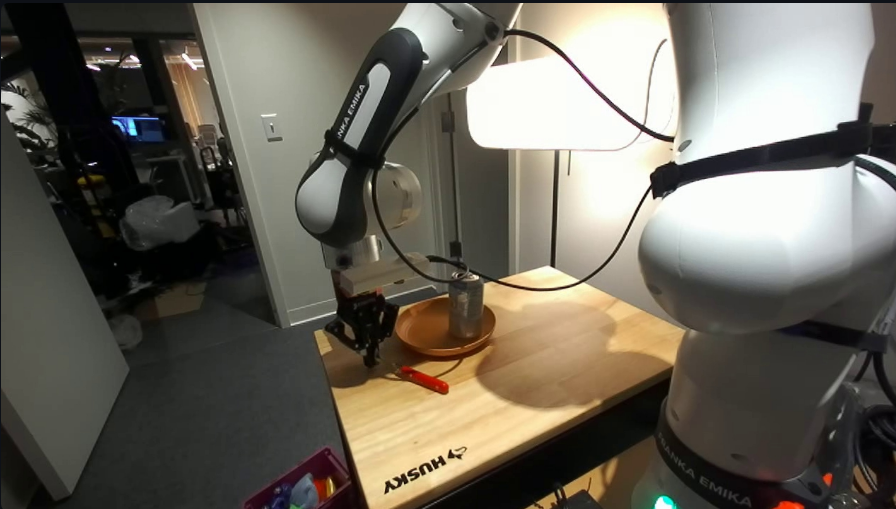}
            \caption*{\texttt{end\_second}: 9.5}
        \end{minipage}
        \caption{Grab the spoon}
        \label{fig:step3}
    \end{subfigure}
    \begin{subfigure}[b]{0.45\columnwidth}
        \centering
        \begin{minipage}[b]{0.45\linewidth}
            \centering
            \includegraphics[width=\linewidth]{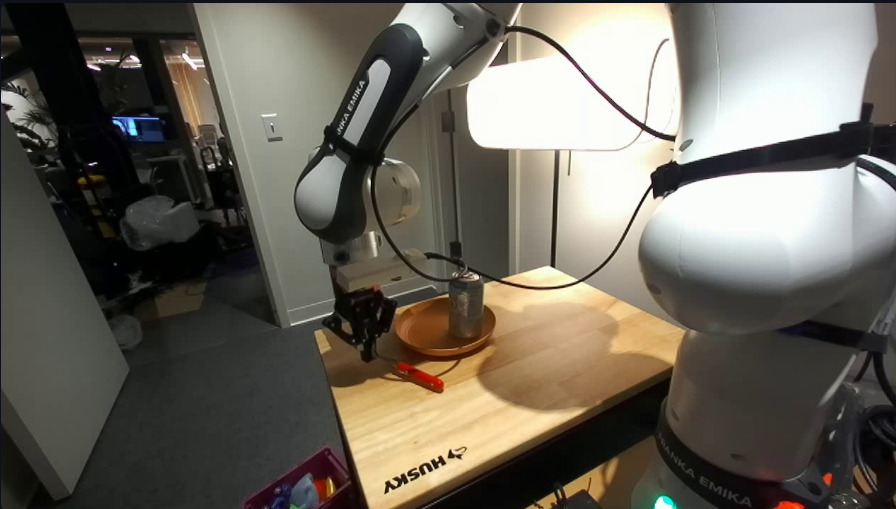}
            \caption*{\texttt{start\_second}: 9.6}
        \end{minipage}
        \begin{minipage}[b]{0.45\linewidth}
            \centering
            \includegraphics[width=\linewidth]{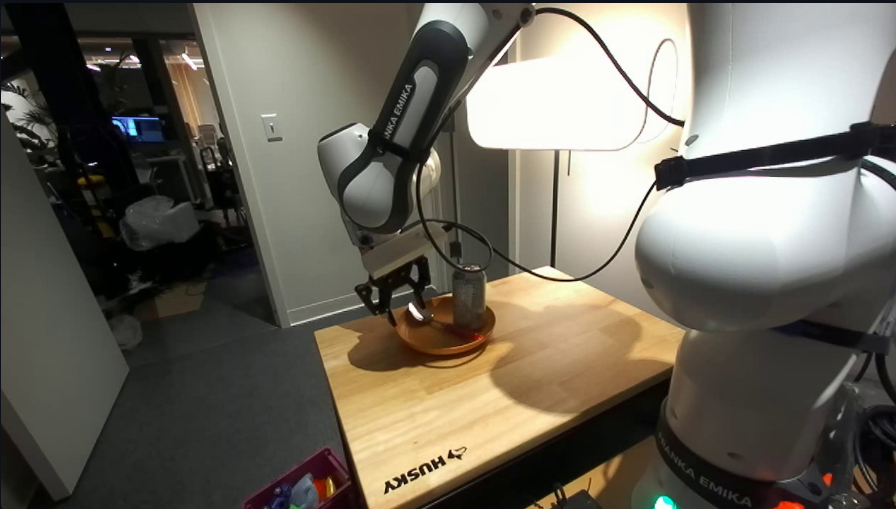}
            \caption*{\texttt{end\_second}: 11.4}
        \end{minipage}
        \caption{Place the spoon in the plate}
        \label{fig:step4}
    \end{subfigure}

    \caption{Expert demonstration with annotation for the task "Clean the table".}
    \label{fig:annotation_steps}
  \end{center}
\end{figure}

\end{document}